\documentclass[10pt,twocolumn,letterpaper]{article}

\usepackage[pagenumbers]{cvpr} 

\usepackage{graphicx}
\usepackage{amsmath}
\usepackage{amssymb}
\usepackage{booktabs}
\usepackage{multirow} 
\usepackage{rotating}
\usepackage{makecell}
\usepackage{multirow}
\usepackage{arydshln}
\usepackage{algorithm}
\usepackage{algorithmic}
\usepackage{chngcntr}%
\usepackage[pagebackref,breaklinks,colorlinks]{hyperref}
\usepackage[misc]{ifsym}
\usepackage{paralist}
\usepackage[capitalize]{cleveref}
\crefname{section}{Sec.}{Secs.}
\Crefname{section}{Section}{Sections}
\Crefname{table}{Table}{Tables}
\crefname{table}{Tab.}{Tabs.}

\newcommand{\revise}{}
\newcommand{\xia}{}
\newcommand{\jimmy}{}
\newcommand{\abc}{}

\begin{document}

\title{A Lightweight and Detector-free 3D Single Object Tracker on Point Clouds}

\author{
	\small
	\begin{tabular}{c c c c c}                                                          
		  Yan Xia $^{1* \footnotemark[2]}$ &
		Qiangqiang Wu$^{2 \footnotemark[2]}$ &
		Wei Li$^3$ &
		Antoni B. Chan$^2$ &
		Uwe Stilla$^1$ \\                                        
		\multicolumn{5}{c}{$^1$Technical University of Munich $^2$City University of Hong Kong  $^3$ Inceptio X-Lab} \\                                                
		\multicolumn{5}{c}{\{yan.xia,stilla\}@tum.de, \{qiangqwu2-c, abchan\}@cityu.edu.hk, liweimcc@gmail.com} 
	\end{tabular}                                                                       
}

\maketitle
\renewcommand{\thefootnote}{\fnsymbol{footnote}} 
\footnotetext[2]{Equal contribution. * Corresponding author.} 
\begin{abstract}
Recent works on 3D single object tracking treat the task as a target-specific 3D detection task, where an off-the-shelf 3D detector is commonly employed for the tracking. However, it is non-trivial to perform accurate target-specific detection since the point cloud of objects in raw LiDAR scans is usually sparse and incomplete. 
In this paper, we address this issue by explicitly leveraging temporal motion cues and propose DMT, a Detector-free Motion-prediction-based 3D Tracking network that completely removes the usage of complicated 3D detectors and is lighter, faster, and more accurate than previous trackers. Specifically, the motion prediction module is first introduced to estimate a potential target center of the current frame in a point-cloud-free manner. Then, an explicit voting module is proposed to directly regress the 3D box from the estimated target center. Extensive experiments on KITTI and NuScenes datasets demonstrate that our
DMT can still achieve better performance ($\sim$10\%  improvement over the NuScenes dataset) and a faster tracking speed (i.e., 72 FPS) than state-of-the-art approaches without applying any complicated 3D detectors.
Our code is released at \url{https://github.com/jimmy-dq/DMT}.
\end{abstract}

\section{Introduction}
Single object tracking (SOT) is a key task in the field of computer vision, which has wide downstream applications in outdoor and indoor scenarios, ranging from autonomous driving \cite{luo2018fast, kiran2021deep}, robot vision \cite{machida2012human, comport2004robust, stoiber2022iterative, stoiber2022srt3d}, and intelligent transportation systems~\cite{zheng2018robust}.
For example, an autonomous pedestrian-following robot should accurately track its master for efficient crowd-following control. Another example is autonomous landing by unmanned aerial vehicles, in which the drone must track the target and know the exact distance and pose of the target in order to land safely~\cite{jiayao2022real}.
In indoor environments, tracking methods~\cite{stoiber2022iterative, stoiber2022srt3d, merrill2022symmetry} can provide the six-degrees-of-freedom (6DoF) pose of an object for robust robotics manipulation.
Given an initial bounding box of a template object in the first frame from images or LiDAR scans, the aim of SOT is to estimate its location by identifying the trajectory across all frames. 
In the past decade, a variety of image-based trackers (e.g., Siamese neural networks~\cite{bertinetto2016fully}) have shown promising performance in the 2D tracking community.
However, the performance of image-based methods often suffers in degraded situations, e.g., when facing drastic lighting changes \cite{shan2021ptt, qi2020p2b}.
As a possible remedy, 3D point clouds collected from LiDAR provide detailed depth and geometric information, which is inherently invariant to lighting changes \cite{xia2021soe}, making it more robust when tracking across frames taken from different illumination environments.

The main challenges of learning-based approaches for 3D SOT trackers are four-fold: 1) a point cloud is structurally unordered compared with images, and thus the network must be permutation-invariant \cite{shan2021ptt}; 2) a point cloud is incomplete because of occlusion or self-occlusion, and thus the network must be insensitive to different resolutions of input point clouds \cite{xia2021vpc}; 3) the scanned point clouds of different objects might have quite similar shapes \cite{zheng2021box}, and thus the network must be insensitive to shape ambiguities; 4) a point cloud has an unstructured nature and thus applying the convolutional operation is difficult ~\cite{xia2021asfm}.

In 3D SOT, the typical solutions follow a Siamese network-based methodology, i.e., comparing the feature similarity between some search regions and the template object. SC3D \cite{giancola2019leveraging} is a pioneering 3D tracker, which first enriches geometric features from sparse point clouds using a shape completion network \cite{achlioptas2018learning}, and then executes template matching with target proposals generated by Kalman filtering \cite{ristic2003beyond}. However, SC3D is not an end-to-end network and also cannot meet the real-time requirement. To address these concerns, P2B \cite{qi2020p2b} first calculates the point-based correlation between the template and the search area, and then applies a Siamese region proposal network (RPN) \cite{li2018high} to detect the final target proposal. Following this, BAT \cite{zheng2021box} explores the free box information to enhance the target-specific search feature.  MLVSNet~\cite{wang2021mlvsnet} proposes performing voting on multi-level features to get more vote centers. With breakthroughs in transformer-based vision methods, the authors of PTT \cite{shan2021ptt} introduce a transformer module to further refine the target-specific search feature. These methods all use historical information to decide the search area, sample seeds in an implicit strategy, and then apply the RPN module (VoteNet \cite{qi2019deep}) to detect the target in the search space.
Although this improves search results, the usage of the RPN module is still complicated and burdensome on the whole. 
Furthermore, the previous 3D trackers ignore one key point: \textit{the coarse target center in the current frame can be directly predicted in a point-cloud-free way by explicitly exploring the historical information. The predicted center can further serve as strong prior knowledge for the final 3D target box prediction.}

To fully utilize this prior knowledge, we propose a novel lightweight and detector-free 3D single object tracking network named DMT (Detector-free Motion prediction-based 3D Tracking). Specifically, we first develop a motion prediction module to estimate the 3D coordinates of a potential target center in the current frame using previous frames. Although the estimated center is coarse, it can provide strong prior information as guidance. Thus, we further design an explicit voting layer only consisting of several multi-layer perceptron (MLP) layers to refine the target center with the desired position and rotation.

To summarize, the main contributions of our work are:
\begin{itemize}
    \item To the best of our knowledge, we are the first to completely remove the usage of complicated 3D detectors or proposal generation in 3D single object trackers. We demonstrate that object motion is a useful cue in 3D SOT, which permits less complex tracking models while still achieving state-of-the-art performance. Our method can serve as a simple yet strong baseline in the 3D SOT community.
    \item We propose a new lightweight and detector-free 3D single object tracking network based on motion prediction, called DMT, and purely applies to point clouds. With the guidance of center priors, an explicit voting module only consisting of several MLP layers is designed to generate accurate 3D positions and the rotation in the \textit{X-Y} plane.
    \item We conduct experiments on the KITTI~\cite{geiger2012we} and NuScenes~\cite{caesar2020nuscenes} benchmark datasets to demonstrate the superiority of DMT over other state-of-the-art 3D SOT methods. 
    Notably, the performance on the NuScenes datasets achieves a $\sim$10\% improvement on average, while running faster and lighter than the previous state-of-the-art methods.
\end{itemize}

\section{Related work}
The goal of object tracking is to locate the object in successive frames using raw data collected from various sensors, which can be 2D images or 3D point clouds.
Numerous methods for tracking objects in 2D or 3D spaces have been developed, which are divided into two categories based on the different data.

{\bf 2D single object tracking.} 2D SOT is a basic computer vision task with a long history spanning decades. The representative deep tracking framework is built on deep Siamese networks \cite{SiamFC}. The pioneering work is SiamFC \cite{SiamFC}, which treats visual tracking as a general template-matching problem and performs favorably in terms of both tracking performance and speed. Based on SiamFC, many improvements have been proposed. SiamDW \cite{siamdw} adopts very deep neural networks (e.g., ResNet \cite{resnet}) as the backbone for Siamese tracking. To handle large-scale appearance variations, SiamRPN \cite{siamrpn} and SiamRPN++ \cite{SiamRPN_plus} employ region proposal networks (RPNs) for scale regression. In addition, much effort is being made to build a robust target appearance model, including UpdateNet \cite{updatenet}, MemTrack \cite{memtrack}, and DSiam \cite{dsiam}. 
Kim \etal~\cite{kim2021discriminative}  presents a strong discriminative appearance model via a novel pooling module.
Recent progress on 3D SOT (e.g., P2B \cite{qi2020p2b} and BAT \cite{zheng2021box}) follows a bounding box regression-based framework, which is mainly inspired by the 2D tracker SiamRPN. However, the data source in 3D tracking is totally different from the images used in 2D tracking. Directly regressing target bounding boxes is still limited when the scanned point clouds are sparse. In this work, we alleviate this problem by incorporating temporal and spatial tracking information for bounding box regression. 

Motion prediction has also been well explored in 2D 
object tracking in videos. However, 2D motion prediction is generally unreliable due to the scale changes,  perspective effects, and inconsistent motion caused by viewing a 2D projection of an object moving in a 3D scene.
 Indeed, most modern deep trackers use a simple learning-free motion prior (e.g., cosine window in SiamFC), and rely on the more reliable 2D appearance features.
 There are a few 2D trackers that use the motion module 
 to \emph{assist} with object detection, e.g., motion-conditioned detection~\cite{hu2019joint, zhou2020tracking, sun2021you} for associating objects in consecutive frames and motion-guided multiple proposal generation~\cite{wang2020motion, liu2020object}. Notably, these trackers still require an object detector module (e.g., RPN) performing on a per-frame basis.
In contrast to 2D SOT,  we show that motion cues in 3D point cloud tracking are more reliable and can be exploited to build lightweight trackers that do not use complex detectors, while still achieving state-of-the-art performance.

{\bf 3D single object tracking.} Early 3D SOT methods \cite{asvadi20163d, bibi20163d, kart2018make, kart2019object} generally rely on the RGB-D information and employ the 2D Siamese tracking architecture. Though these methods are effective in certain situations, they do not fully explore 3D geometric clues. SC3D \cite{giancola2019leveraging} is a pioneering work for point-cloud-based tracking, which regularizes the latent spaces of the template point cloud and search candidates using a shape completion network. However, this method is time-consuming since it uses Kalman filtering for the target proposal generation. Moreover, it ignores the local geometric information of each target proposal. PSN \cite{cui2020point} leverages 3D Siamese network for single-person tracking. However, it cannot predict the orientation and size of the target. F-Siamese tracker \cite{zou2020f} explores RGB images to produce 2D region proposals to reduce the 3D point cloud searching space. However, its performance depends more on the 2D tracker. 3DSiamRPN \cite{fang20203d} combines a 3D Siamese network and a 3D RPN to track a single object, but the one stage RPN network limits its performance. P2B \cite{qi2020p2b} fuses the target object information into 3D search space and then adopts a state-of-the-art object detection network (VoteNet) to detect the target. Following this, BAT \cite{zheng2021box} proposes adding the bounding box information provided in the first frame as an additional cue. MLVSNet~\cite{wang2021mlvsnet} performs Hough voting on multi-level features to get more vote centers. PTT \cite{shan2021ptt} introduces the transformer architecture to enhance the target-specific feature extracted in P2B. However, these methods all use the RPN to regress the bounding box of the target, which is inspired by their 2D SOT counterparts \cite{siamrpn,siamdw,SiamRPN_plus}. In this paper, we show that complex detectors can be removed from 3D SOT by better leveraging more reliable 3D motion prediction, and still achieving state-of-the-art performance. 

\section{Problem statement}
Let \textit{${B_{init}} = {\left \{x, y, z, h, w, l, \theta \right \}}$} be a known 3D bounding box of the object in the first frame, where $(x, y, z)$ are the center coordinates of the 3D bounding box, $(h, w, l)$ are the height, width, and length respectively, and $\theta$ is the orientation of the bounding box. Further, let $ {Q = \{ Q_{i} \}_{i=1}^M}$ be a query point cloud created by cropping and centering the object in the first frame with \textit{${B_{init}}$}. $Q_{i}$ is a 3D point in the $Q$.
We define the single object tracking task as locating the same object in the search point cloud ${P =  \{ P_{i} \}_{i=1}^N}$ given the \textit{${B_{init}}$} frame by frame.  $M$ and $N$ are the number of points in the query point cloud and search point cloud, respectively. 
Formally, previous state-of-the-art 3D single object trackers \cite{qi2020p2b,zheng2021box} can be formulated as:
\begin{equation} 
\begin{aligned}
Tracker\left ( Q, P, B_{init} \right ) \rightarrow  ( \hat{x},\hat{y} , \hat{z}, \hat{\theta }),
\end{aligned}
\label{Eq:tracker}
\end{equation}
\noindent 
where $Q \in \mathbb{R}^{M\times 3}$, $P \in \mathbb{R}^{N\times 3}$, and $B_{init} \in \mathbb{R}^{7}$. Notably, we only predict the center coordinates and orientation $( \hat{x},\hat{y} , \hat{z}, \hat{\theta } )$ of the target since the height, width, and length of the object are assumed to be the same in other frames.

Previous trackers employ off-the-shelf detectors on scanned point clouds for target detection. They may easily drift when the point clouds are relatively sparse or incomplete. In this paper, we propose predicting the potential target center in a point-cloud-free way, that fully explicitly leverages motion cues from previous target states $S_{prev} = \{ S_{1}, S_{2},\cdots , S_{t-1}  \}$, where the state $S_{t}$ is the predicted center coordinates in the $t$-th frame. The whole process is formulated as:
\begin{equation} 
\begin{aligned}
Tracker\left ( Q, P, B_{init}, \mathcal{M}(S_{prev})  \right ) \rightarrow  ( \hat{x},\hat{y} , \hat{z}, \hat{\theta } ),
\end{aligned}
\label{Eq:tracker}
\end{equation}
\noindent 
where \textit{$\mathcal{M}(\cdot)$} is a motion prediction function that estimates a potential target center in the current frame based on previous target states.  

\section{Methodology}

\begin{figure*}[ht!]
\centering
\includegraphics[width=0.95\linewidth]{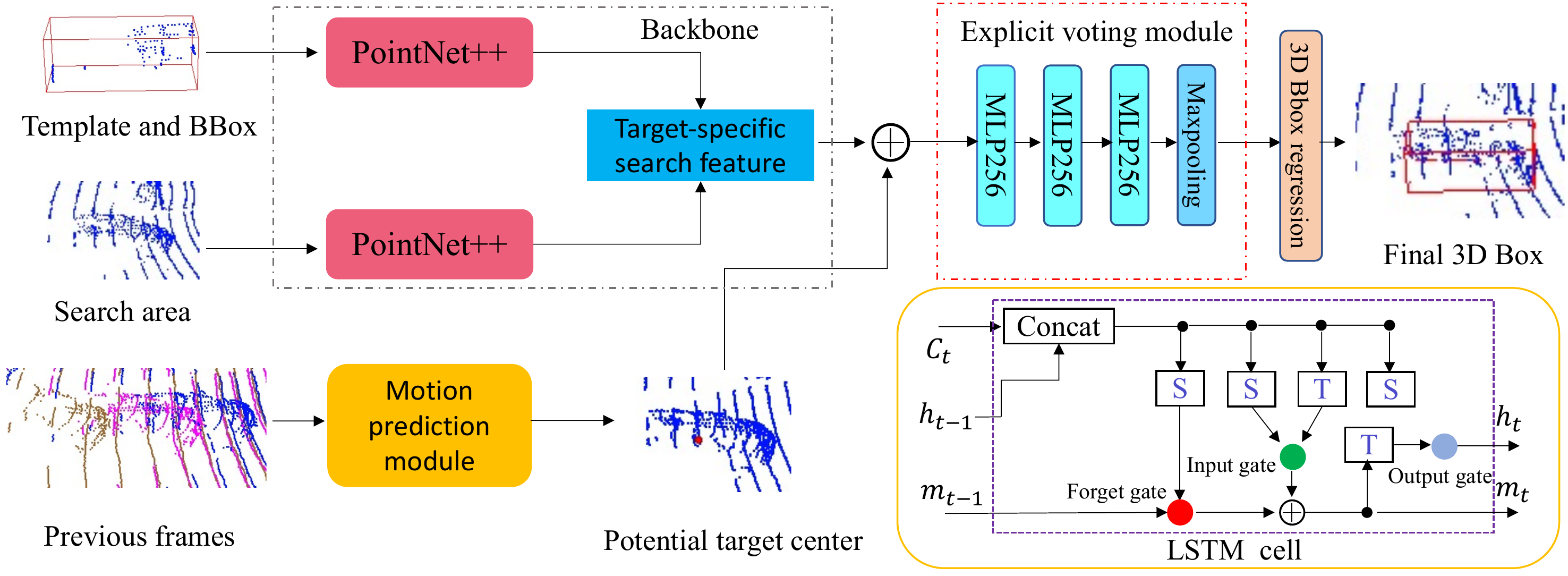}    
\caption{Overview of DMT. The backbone network first extracts the target-specific features from the template and search area points following  \cite{zheng2021box}. Then the motion prediction module (MPM) estimates the 3D coordinates of a potential target center. Next, the explicit voting module refines the target-specific search feature extracted by the backbone to
the coarse predicted center. Finally, a 3D bounding box prediction head regresses the target location. One example of the MPM is an LSTM (lower right corner).}
\label{fig:network} 
\end{figure*}

The overall network architecture of our DMT is shown in Fig.~\ref{fig:network}.
Given the query and search point cloud with coordinates denoted as $Q$ and $P$, and an initial bounding box ${B_{init}}$, we first use the backbone to extract target-specific features following \cite{zheng2021box}, as introduced in Section~\ref{sec: backbone}. 
Unlike previous studies, we propose a motion prediction module to estimate a potential target center in the current frame based on previous target states \textit{$S_{prev}$}, with details described in Section~\ref{sec: motion_prediction}. Afterward, an explicit voting module is adopted to modify the coordinates of the coarse predicted center and predict the orientation in Section~\ref{sec: explicit voting module}. 
The loss function is presented in Section~\ref{sec: loss function}.
The training strategy and implementation details are explained in Section~\ref{sec:  Implementation}.
To highlight the simplicity of our method, we also sketch the detailed flow in Algorithm~\ref{alg1}. 

\begin{algorithm}[t!]
\small
	\renewcommand{\algorithmicrequire}{\textbf{Input:}}
	\renewcommand{\algorithmicensure}{\textbf{Output:}}
	\begin{algorithmic}[1]
		\REQUIRE Points $Q$ in query, points $P$ in search area, an initial bounding box ${B_{init}}$, previous target states $S_{prev}$, and target-specific search feature $f$.
		\STATE  {\bf Potential target center generation.} Given $S_{prev}$, predict a coarse target center $C_{coarse}$ in the current frame using a motion prediction module.
		\STATE {\bf Explicit voting.} Feed $f$ and $C_{coarse}$ into an explicit voting module to estimate the target-specific point feature $\hat{f}$ of the target center.
		\STATE {\bf Final box  regression.} Regress the 3D bounding box of the target based on $\hat{f}$ using a prediction head.
		\ENSURE  The 3D bounding box of the target.
	\end{algorithmic}  
	\caption{The Workflow of DMT}
	\label{alg1}
\end{algorithm}

\subsection{Backbone} 
\label{sec: backbone}
\begin{figure}[ht!]
\vspace{-0.5cm}
\centering
\includegraphics[width=0.9\linewidth]{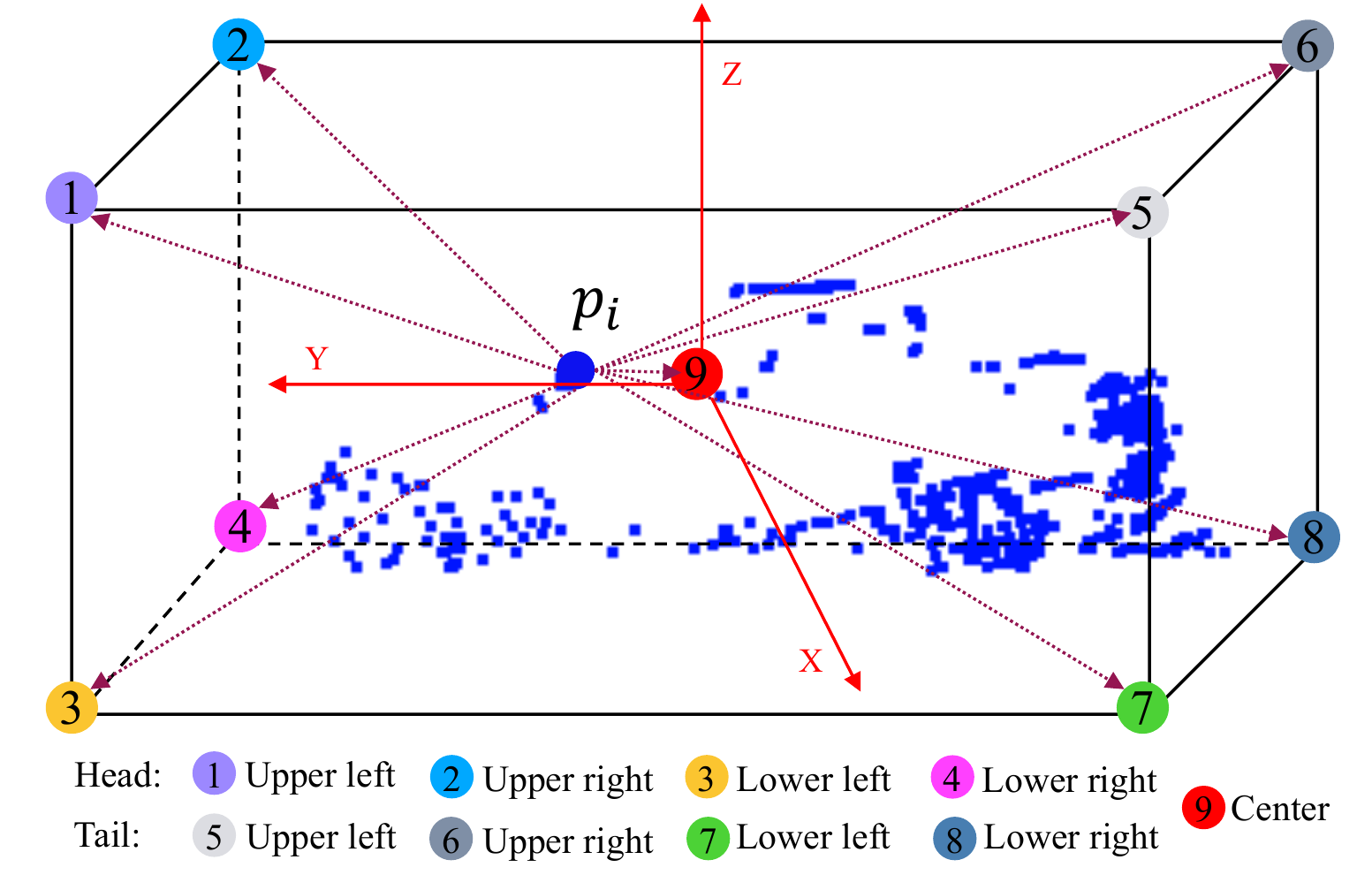}   
\caption{Illustration of a BoxCloud. The BoxCloud is a set of 9D coordinates. Each 9D coordinate consists of the distances from one point to eight corners and the center of its 3D bounding box.}
\label{fig:boxcloud} 
\end{figure}

\begin{figure}[ht!]
\centering
\includegraphics[width=1.0\linewidth]{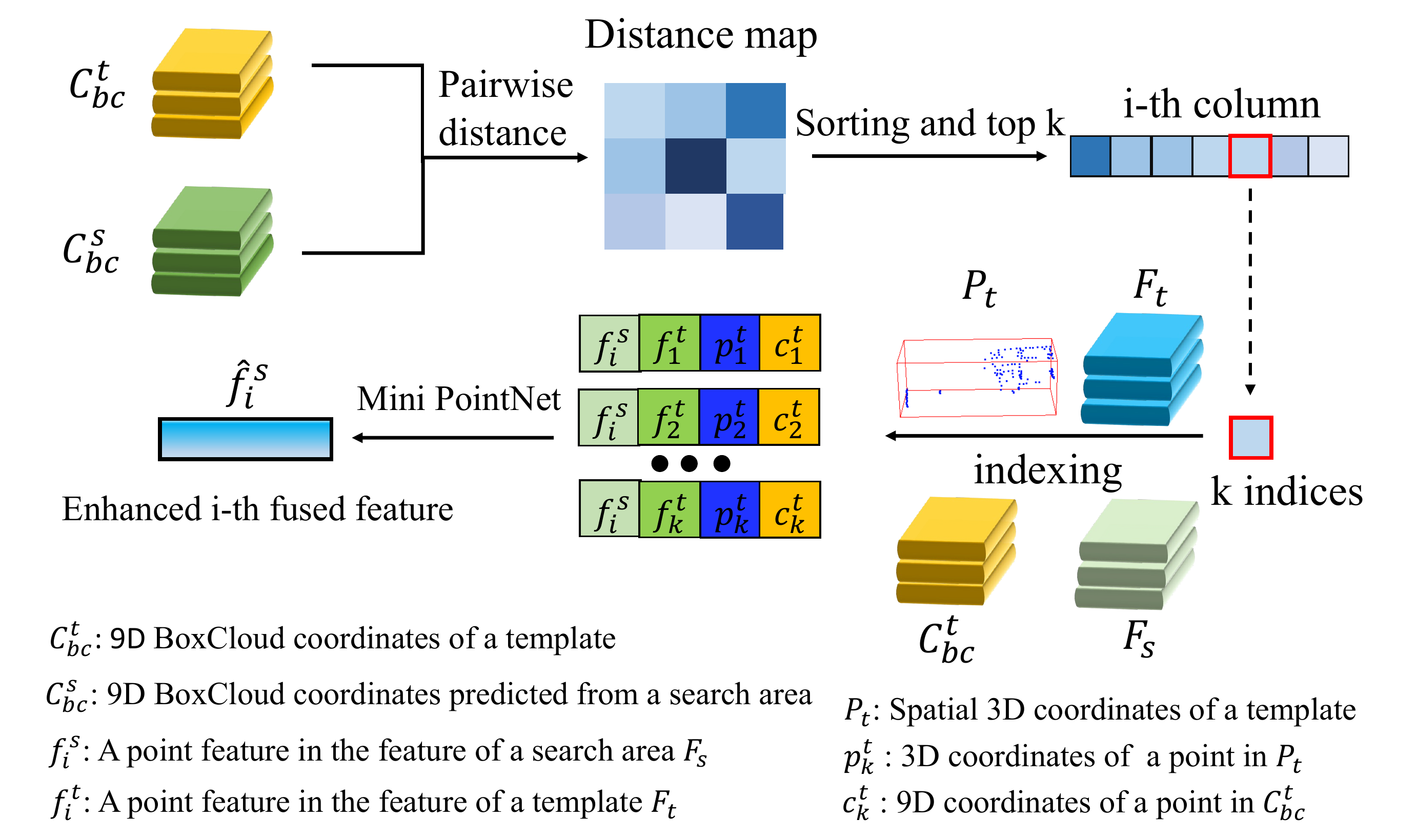}   
\vspace{-0.5cm}
\caption{\revise{The workflow of the box-aware feature fusion (BAFF) module. ${C_{bc}^{s}}$ is the 9D BoxCloud coordinates, predicted from each search point feature $f_{i}^{s}$ via MLP. ${C_{bc}^{t}}$, ${F_{t}}$, $P_{t}$ are the 9D BoxCloud coordinates, the features, and the spatial 3D coordinates of a template, respectively. The module first generates the distance map between the BoxCloud ${C_{bc}^{s}}$ and ${C_{bc}^{t}}$ to retrieve the top-$k$ nearest neighbors with respect to each point in the search area. Then, a mini-PointNet is adopted to generate $\hat{f_{i}^{s}}$ by aggregating the neighbors' features.}}
\label{fig:BAFF} 
\vspace{-0.25cm}
\end{figure}
The aim of the backbone network is to generate an enhanced target-specific search feature by fusing the template's target information into the search area points. We adopt the box-aware feature fusion (BAFF) module in \cite{zheng2021box} as our backbone\footnote{Our framework is not restricted to BAFF, and any suitable backbone could be used.}, as shown in Fig.~\ref{fig:BAFF}. The template and search area are first fed into PointNet++ \cite{qi2017pointnet++} to obtain their features. Then the BAFF module help augment the search area with target-specific features, which includes BoxCloud \cite{zheng2021box} comparison and feature aggregation sub-modules.
\revise{A BoxCloud is defined by the point-to-box relation between an object point cloud and its 3D bounding box. For each point $p_{i}$ in this point cloud, nine Euclidean distances from the $p_{i}$  to each of the eight corners and the center of the bounding box are calculated. 
As shown in Fig.~\ref{fig:boxcloud}, every point is represented by a
9D vector $c_{i}$. Formally, a BoxCloud $C_{bc}$ can be formulated as follows:
\begin{equation}
C_{bc}=\left\{c_i \in \mathbb{R}^9 \mid c_{i j}=\left\|p_i-q_j\right\|_2, \quad \forall j \in[1,9]\right\}_{i=1}^N,
\end{equation}
\noindent
where $q_{j(j \neq 9)}$ is the $j$-th corner and $j_{9}$ is the center of the bounding box.}


{\bf BoxCloud comparison.} \revise{Given the feature of a search area  ${F_{s} = \{ f_{i}^{s} \}_{i=1}^{M_{1}}}$ obtained by PointNet++, we predict the 9D BoxCloud coordinates ${C_{bc}^{s} = \{ c_{i}^{s} \in \mathbb{R}^9 \}_{i=1}^{M_{1}}}$ from each point feature $f_{i}^{s}$ via MLP, where $M_{1}$ is the number of points in $C_{bc}^{s}$.} The prediction is supervised by a BoxCloud loss, presented in Sec. \ref{sec: loss function}. 
Then we compare the pairwise distance between the predicted ${C_{bc}^{s}}$ and the BoxCloud ${C_{bc}^{t} = \{ c_{i}^{t} \}_{i=1}^{M_{2}}}$ of the template, as shown in Fig.~\ref{fig:BAFF}, where $M_{2}$ is the number of points in $C_{bc}^{t}$. Following \cite{zheng2021box}, we adopt the simple $l_{2}$ distance as the distance metric. After obtaining the distance map, we sort and select the top $k$ most similar template points for each point in the search area. The $i$-th column of the distance map in Fig.~\ref{fig:BAFF} represents the indices of the $k$ nearest neighbors of the $i$-th search point $p_{i}^{s}$. 

{\bf Feature aggregation.} \revise{After getting the top $k$ template features, we hope to fuse them into the search area. 
Considering the feature of a template ${F_{t}}$ extracted from PointNet++, the corresponding spatial 3D coordinates} $P_{t}$, and 9D BoxCloud coordinate $C_{bc}^{t}$ of the template points, we construct more informative $k$ tuples $\left \{ \left [  f_{j}^{t},p_{j}^{t},c_{j}^{t},f_{i}^{s}\right ],\forall j=1,\cdots ,k\right \}$. Finally, a mini-PointNet is used to obtain the aggregated feature of the search point from these pairs, which can be formulated as follows:
\begin{equation} 
\begin{aligned}
\hat{f_{i}^{s}}=G\odot \left \{ MLP(\left [  f_{j}^{t},p_{j}^{t},c_{j}^{t},f_{i}^{s}\right) \right \}_{j=1}^{k}),
\end{aligned}
\label{Eq:feature aggregation}
\end{equation}
\noindent 
where $G\odot$ is a max-pooling operation. Finally, we can get the effective target-specific search feature $\hat{F_{s}} =  \{  \hat{f_{i}^{s}} \}_{i=1}^{M_{2}}$.

\subsection{Motion prediction module} 
\label{sec: motion_prediction}
The previous end-to-end 3D SOT methods \cite{qi2020p2b, zheng2021box, shan2021ptt} heavily rely on point cloud features 
for target object detection. However, 
erroneous detection may occur when the point cloud of the target is incomplete~\cite{giancola2019leveraging}. To alleviate this, we propose explicitly leveraging spatio-temporal information for 3D SOT. Specifically, we introduce a motion prediction module (MPM) $\mathcal{M}$ based on previous target states (i.e., predicted 3D target center coordinates in the previous frames) to predict a coarse target center in the current frame. Suppose that we have a tracklet $\{(x_{i}, y_{i}, z_{i})\}_{i=1}^{t}$ in the previous $t$ frames; the prediction of the target center location in the next $(t+1)$-th frame is formulated as:  
\begin{align}
\label{mp}
(\hat{x}_{t+1}, \hat{y}_{t+1}, \hat{z}_{t+1}) = \mathcal{M}(\{(x_{i}, y_{i}, z_{i})\}_{i=1}^{t}).
\end{align}
In our general design, common regression or prediction models can be employed as our MPMs for effective target center prediction. Here we introduce several simple yet effective MPMs.

{\bf Constant velocity model.} The constant velocity model assumes that the target acceleration in the current frame is 0, and the velocity of the target in the current frame should be equal to the velocity in the last frame. Given the target locations in the $(t-1)$ and $t$-th frames $\{(x_{i}, y_{i}, z_{i})\}_{i=t-1}^{t}$, the predicted target center coordinates in the $(t+1)$-th frame are calculated as $(2x_{t}-x_{t-1}, 2 y_{t}-y_{t-1}, 2z_{t}-z_{t-1})$. Despite the simplicity of this model, we find it also works very well in our DMT.

{\bf Sequence-to-sequence prediction model.} The goal of our MPM is to predict 3D coordinates based on previously estimated $t$ target coordinates, which is actually a sequence-to-sequence prediction task. A long short-term memory (LSTM) network \cite{hochreiter1997long} is a typical sequence-to-sequence prediction model that has been widely used in various sequence prediction tasks. In this paper, we choose a multi-layer LSTM since this naive LSTM model can better validate the effectiveness of our proposed tracking method. The conventional LSTM cell is shown in Fig.~\ref{fig:network} (bottom right). More details 
can be found in \cite{hochreiter1997long}. In the implementation, we select the center coordinates of the 10 consecutive frames from the times $t-10$ to $t$ to predict potential target center coordinates in the $(t+1)$-th frame.
 In the training stage, we prepare multiple training tracklets generated from the KITTI and NuScenes datasets to train the LSTM. In online tracking, we directly use the offline trained LSTM network for motion prediction without further updating.

{\bf Regression model.} Traditional learning-based regression models can also be employed as MPMs. In this paper, we try several basic regression models, including linear regression, ridge regression, Gaussian processor regression, and RANSAC regression. The training for the above models is similar to the LSTM-based MPM, i.e., using the generated tracklet training data for training in an offline manner.

The above basic MPMs can roughly predict the potential target center coordinates based on the previous states. The prediction is not always reliable since the previous target states may be noisy (i.e., the predicted target center does not match the ground truth), or the target changes position in an unexpected way.  To alleviate this problem, we propose a lightweight explicit voting module to further refine the MPM prediction.

\begin{figure*}[ht!]
\centering
\includegraphics[width=0.9\linewidth]{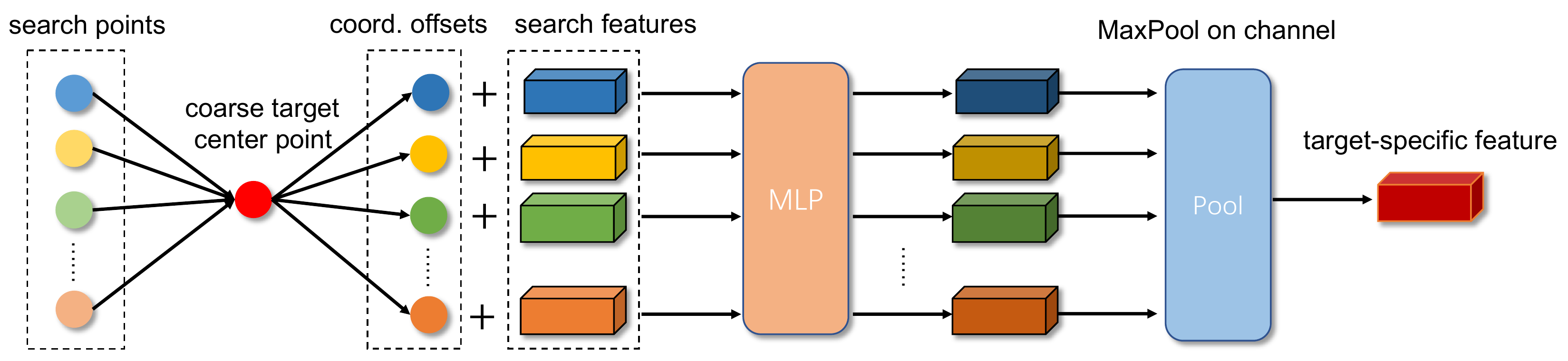}    
\caption{The overall pipeline of the explicit voting module (EVM). 
Our EVM first calculates the coordinate offsets between each search point and the coarse predicted target center. Then the offsets are jointly concatenated with the search features for feature modeling via an MLP. Finally, a permutation-invariant max pooling layer is applied to obtain the target-specific feature of the predicted target center point for the final 3D box prediction.}
\label{fig:epv} 
\end{figure*}
\subsection{Explicit voting module}
\label{sec: explicit voting module}

Before going into the details of our proposed explicit voting module (EVM), we give a short review of the RPN module (VoteNet) used by previous trackers \cite{qi2020p2b, zheng2021box, wang2021mlvsnet, shan2021ptt}. The architecture of VoteNet includes two aspects: 1)  Hough voting to convert the search area seeds into possible target centers; and 2) clustering neighboring potential target centers to obtain the final target center. To generate the potential target centers, VoteNet estimates the coordinate offsets between each search seed and ground-truth target center, which aims to push the predicted possible target centers and ground-truth target center to be as close as possible. In our DMT, the above two steps can be removed since the coarse target center location in the current frame is provided by our MPM, which makes our method simpler and lighter.

The overall pipeline of our proposed explicit voting module is shown in Fig. \ref{fig:epv}. As can be seen, after obtaining the coarse target center coordinates $(\hat{x}_{t+1}, \hat{y}_{t+1}, \hat{z}_{t+1})$ estimated by the MPM and the target-specific search feature, the goal of our EVM is to  estimate effective features at $(\hat{x}_{t+1}, \hat{y}_{t+1}, \hat{z}_{t+1})$. In the design of the EVM, 
we use coordinate offsets as explicit voting signals to estimate the target center feature. Specifically,  we first calculate the coordinate offset between the estimated target center and each search point. We then concatenate the coordinate offset with the search point feature to obtain a candidate voting feature $f \in \mathbb{R}^{C+3}$, where $C$ denotes the feature dimension. Suppose there are $N$ search points with $N$ corresponding candidate voting features $\{f_{i}\}_{i=1}^{N}$.  The explicit target coordinate voting is formulated as:
\begin{align}
\label{exp}
\bar{f}_{i} = \mathrm{MLP}(f_{i}), \quad
\hat{f} = \mathrm{MaxPool}(\{\bar{f}_{i}\}_{i=1}^{N}),
\end{align}
where \jimmy{$\bar{f}_{i}  \in \mathbb{R}^{C}$}, and $\hat{f} \in \mathbb{R}^{C}$ are the final estimated target-specific feature \jimmy{at} the estimated target center, \jimmy{which is obtained by applying the max pooling operation on the channel dimension of each feature vector in $\{\bar{f}_{i}\}_{i=1}^{N}$}. The estimated feature $\hat{f}$ is finally fed into a prediction head (i.e., MLP) to regress the bounding box of the target.

In the training stage, given a ground-truth target center location in a frame, we randomly sample diverse points around the ground-truth center. For stable training, the maximum distance between the sampled points and the ground-truth center should not be too large, and here we set it to 0.75 meters. During training, our EVM learns to estimate target-specific features of the sampled points that are effective for predicting the final bounding box. Note that the diverse sampled points can effectively mimic the noisy predictions of MPM, which makes our DMT less sensitive to noise in the predicted target track.

\subsection{Loss function}
\label{sec: loss function}
Following \cite{zheng2021box}, our training loss includes three components: classification loss,  box-cloud loss, and regression box loss. The first two losses enhance the target-specific feature extracted by the backbone,  while the latter supervises the estimated 3D bounding box.
 In addition, we add a \revise{motion prediction loss} to train the MPM 
(except for the constant velocity model).

{\bf Point-wise classification loss.} Following \cite{qi2020p2b}, we note that only search points located on the surface of a ground-truth target are useful in the EVM, and thus labeled as positives, while all others are negatives. Therefore, a standard binary cross entropy loss $L_{cla}$ is adopted to classify the search points. 

{\bf BoxCloud loss.} The BoxCloud features~\cite{zheng2021box} in the search area are unknown in the inference stage, so we need to predict the 9D BoxCloud coordinate $C_{bc}$ in the search area, which is supervised by a smooth-L1 regression loss. 
\begin{equation} 
\begin{aligned}
\mathcal{L}_{bc} = \frac{1}{\sum _{i}\xia{E}_{i}}\sum_{i=1}^{N} \left \|C_{bc}^{i} - \hat{C}_{bc}^{i}   \right \|\cdot E_{i} ,
\end{aligned}
\label{Eq:box-cloud loss}
\end{equation}
\noindent 
where $\hat{C}_{bc}$ are ground-truth BoxCloud coordinates pre-calculated before training. \textit{$E_{i}$} is a binary mask, which indicates whether the \textit{$i$}-th point is inside an object BBox or not.

{\bf 3D box regression loss.} The final result of our network is to predict the 3D box parameters $ C_{bbox} =  \{\hat{x},\hat{y} , \hat{z}, \hat{\theta } \}$. Following previous work, we adopt Huber (smooth-L1 loss) to supervise the regression.
\begin{equation} 
\begin{aligned}
\mathcal{L}_{bbox} = \left \|C_{bbox} - \hat{C}_{bbox}   \right \|,
\end{aligned}
\label{Eq:box regression loss}
\end{equation}
\noindent 
where $\hat{C}_{bbox}$ is the ground-truth bounding box of the target.

{\bf \revise{Motion prediction loss.}} When training an MPM, 
we hope the distance between the predicted center coordinates of the target and the ground truth is as small as possible. In this paper, we use the mean squared error loss  $\mathcal{L}_{v}$ for supervision:
\begin{equation} 
\begin{aligned}
\mathcal{L}_{v} = \big \|C_{cen}^{t+1} - \hat{C}_{cen}^{t+1}   \big \|_{2},
\end{aligned}
\label{Eq:box regression loss}
\end{equation}
\noindent 
where $C_{cen}^{t+1}=(\hat{x}_{t+1}, \hat{y}_{t+1}, \hat{z}_{t+1})$ (see Eq. (\ref{mp})) is the predicted target center coordinates at the  $(t+1)$-th frame, and $\hat{C}_{cen}^{t+1} $ is the corresponding ground-truth coordinates. 

Note that we first train the MPM with $\mathcal{L}_{v}$, and then we use the following combined loss to train the backbone network, EVM,  and the prediction head:
\begin{equation}
\begin{aligned}
L = \alpha L_{cla}+ \beta L_{bc}+\gamma L_{bbox},
\end{aligned}
\label{Eq:overall loss}
\end{equation}
\noindent 
where $ \alpha $, 	$ \beta $, and $ \gamma $ are  hyperparameters to balance their relationship. Here we set $\alpha = 0.2$, $\beta = 1.0$, $\gamma = 0.2$.

\subsection{Implementation details}
\label{sec: Implementation}
We follow previous 3D trackers \cite{qi2020p2b,zheng2021box} to generate templates and search point clouds in both the training and testing stages. To fairly compare with recent trackers equipped with online detectors, we use the same target-specific search feature generation method in BAT \cite{zheng2021box}, which makes the predictions of BAT and our DMT both based on the same augmented search features.

{\bf Search area generation.} In practice, the object movement between two consecutive frames is relatively small, so searching the entire frame for the target is unnecessary. Following \cite{zheng2021box}, we look for the target near the previous object location to generate search areas for training and testing. During both training and testing, templates and their BBoxes are transformed into the object coordinate system before being sent to the model.

{\bf Network architecture.} In the proposed MPM, we use one LSTM layer with 50 hidden units as the motion predictor. The input tracklet length is set to 10, meaning that  we use target states in the previous 10 frames for prediction. The model size of this LSTM model is about 50K, which is extremely light. The EVM is implemented as a three-layer MLP with 256 hidden units, where the first two layers are followed by a 1D batch normalization layer and a ReLU activation layer. We use the same backbone and box prediction head as P2B \cite{qi2020p2b} and BAT \cite{zheng2021box}.

{\bf Training.} In the training stage, we first generate tracklet training data (i.e., each tracklet contains the target center coordinates in every 10 frames and the corresponding ground-truth target center coordinates in the next frame) to train the LSTM network. The batch size is set to the overall dataset size, and the learning rate and training epochs are respectively set to 1e-3 and 8,000. The whole training takes only 28 seconds for the car category of the KITTI dataset, which is efficient. After training the LSTM network in an offline manner, we use it for online testing without further modifications. The proposed DMT is trained for 60 epochs using the Adam optimizer with a batch size of 100. The learning rate is initialized as 1e-3 and decayed by 0.5 in every 5 epochs. 

{\bf Testing.} During testing, we apply the trained DMT to infer the 3D bounding boxes of a given target within tracklets frame by frame. For the current frame, the template is updated by fusing the point clouds in the first given BBox and in the previously estimated BBox. To obtain the search area, \jimmy{we enlarge the previously estimated BBox by 2 meters in the current frame and collect the points within the enlarged BBox.}

\section{Experiments}
In this section, we first describe the experimental settings. Next, we present experiments on the KITTI and NuScenes datasets to demonstrate the efficacy of our lightweight 3D SOT tracker, DMT.
\label{sec:Experiments}
\subsection{Dataset}
The KITTI dataset \cite{geiger2012we} includes raw point clouds scanned by the Velodyne HDL-64E rotating 3D laser scanner and annotations for object instances in the form of 3D bounding boxes. The tailored dataset contains 21 outdoor scenes and 8 categories of targets. Following \cite{qi2020p2b}, we generate tracklets for target instances within all videos and split the KITTI training set into three parts: scenes 00-16, scenes 17-18, and scenes 19-20 for the training, validation, and test sets, respectively,  since the annotations of the test set in KITTI are inaccessible. Furthermore, we also conduct experiments on the more challenging dataset NuScenes~\cite{caesar2020nuscenes}. The NuScenes dataset includes 1000 outdoor scenes and 23 categories of objects with annotated 3D bounding boxes. Specifically, the NuScenes dataset contains 32,302 frames in the car category, which is five times larger than the KITTI dataset. Following \cite{zheng2021box}, the training set of NuScenes is used for training, and the validation set is used for testing.

\begin{figure}[t!]
\centering
\includegraphics[width=1.0\linewidth]{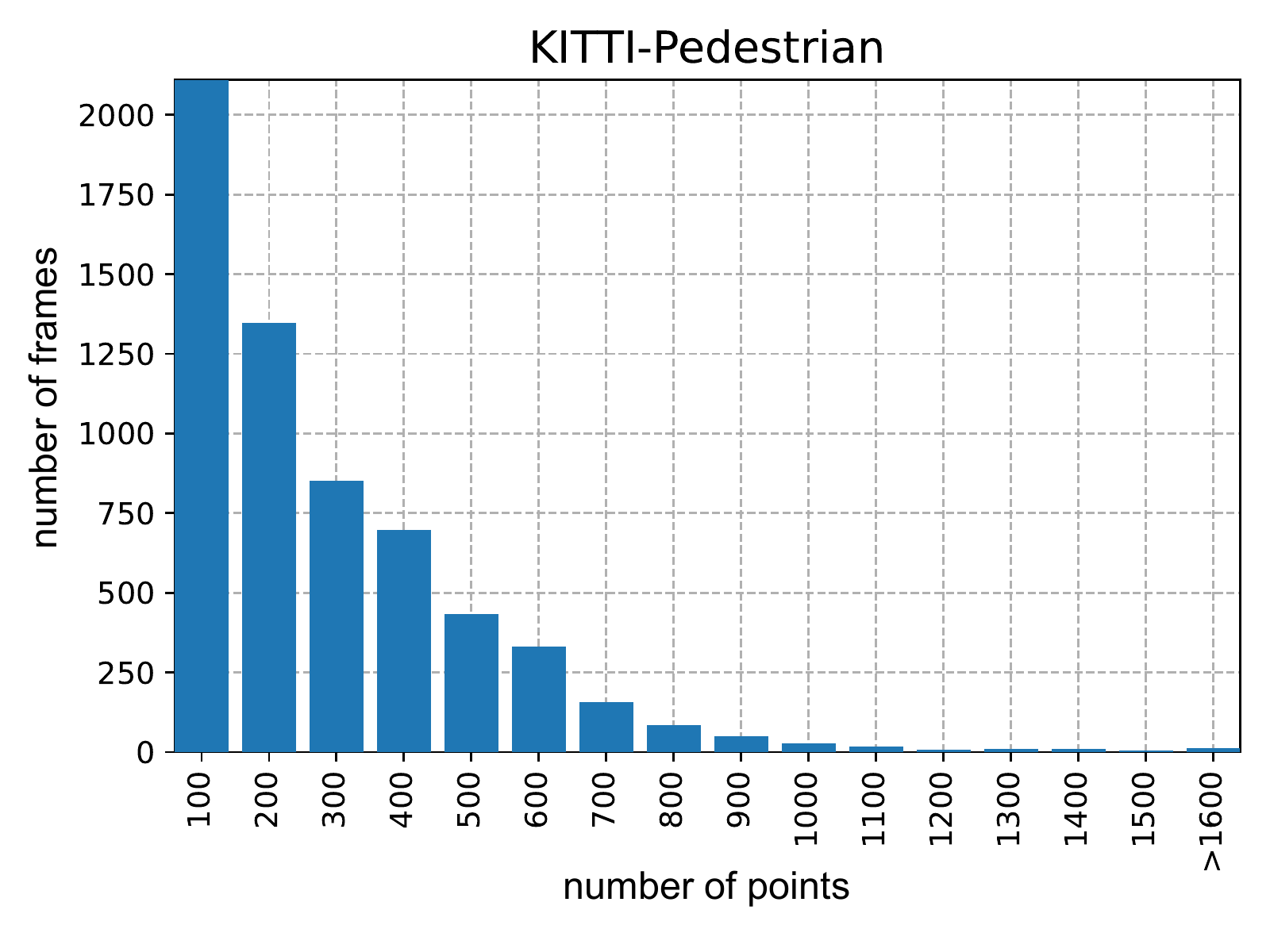}
\caption{Long-tailed distribution of the frame-wise number of points in KITTI-Pedestrian, which shows the sparsity of target points.
}
\label{fig:point_pedes_num} 
\end{figure}

{\bf Sparsity of point clouds.} Although there are (on average) $\sim$120k points in each frame of raw LiDAR data, the points on the target object might be quite sparse due to occlusion and LiDAR defects~\cite{qi2020p2b}. Thus we count the number of points in the pedestrian category of the KITTI benchmark, as shown in Fig.~\ref{fig:point_pedes_num}. About 36\% of pedestrians have fewer than 100 points, and this sparsity introduces great challenges to 3D single object tracking based on point clouds.

\subsection{Evaluation metric}
Following \cite{qi2020p2b, zheng2021box}, we apply One Pass Evaluation (OPE) \cite{wu2013online} to measure the Success and Precision of different approaches. \jimmy{For a predicted bounding box and a ground-truth bounding box, ``Success'' is defined  as the intersection over union (IOU) between them. ``Precision'' is defined as the AUC for the distance error curve from 0 to 2m, which is measured between the centers of the two boxes. The success and precision metrics, respectively, measure the box overlap and center distance error between the predicted bounding box and the ground-truth bounding box.}

\subsection{Comparison with State-of-the-arts}
\label{sec: comparision}
\begin{figure*}[t]
\centering
\includegraphics[width=1.0\linewidth]{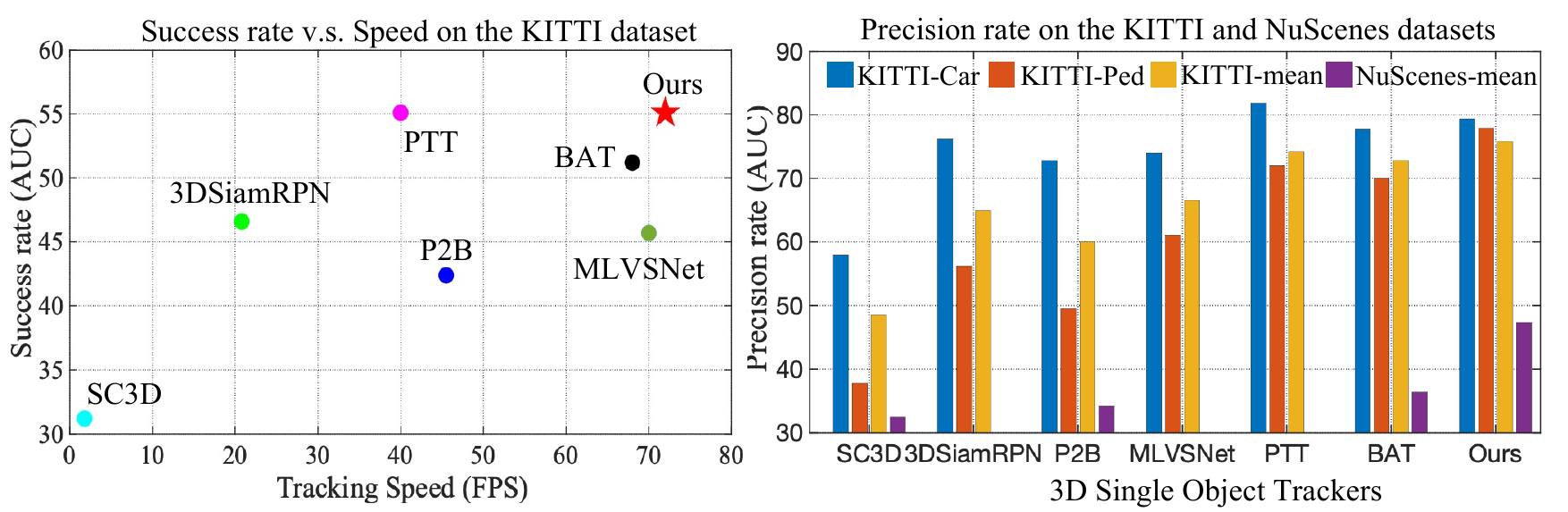}    
\caption{(Left) Tracking accuracy vs.~speed for the Car category of the KITTI benchmark. 
Our DMT outperforms state-of-the-art 3D single-object trackers in terms of both tracking accuracy and speed.
(Right) Precision comparison for KITTI-Car, KITTI-Pedestrian, KITTI-mean, and NuScenes-mean.
} 
\label{fig:cover} 
\vspace{-0.3cm}
\end{figure*}

We compare our network with the state-of-the-art methods: SC3D \cite{achlioptas2018learning}, its follow-up SC3D-RPN \cite{zarzar2019efficient}, FSiamese \cite{zou2020f}, 3DSiamRPN \cite{fang20203d}, P2B \cite{qi2020p2b}, MLVSNet~\cite{wang2021mlvsnet}, PTT~\cite{shan2021ptt}, and BAT~\cite{zheng2021box}. For a fair comparison, we use the same evaluation metrics. In this paper, the default setting of the MPM is an LSTM prediction model. Fig.~\ref{fig:cover} and 
Table \ref{tab: experimental results} show the success and precision of each network on the KITTI and NuScenes datasets. 
The success and precision values for other methods are those reported in their published papers \cite{achlioptas2018learning, zarzar2019efficient, zou2020f, fang20203d, qi2020p2b, wang2021mlvsnet, shan2021ptt, zheng2021box}. We first quantitatively evaluate our network on KITTI, and then extend the comparisons to NuScenes.

\begin{table*}[t]
\small
\caption{Results of the Success and Precision of different 3D trackers with different categories on the KITTI and NuScenes dataset. ‘PED’ REPRESENTS ‘PEDESTRIAN.’}
\label{tab: experimental results}
\centering
\resizebox{1.0\textwidth}{!}{
\begin{tabular}{c|c|ccccc|ccccc} 
\hline
\multicolumn{1}{l|}{\multirow{3}{*}{}} & \multirow{3}{*}{\begin{tabular}[c]{@{}c@{}}Dataset\\Category \\Frame Number\end{tabular}} & \multicolumn{5}{c}{KITTI} & \multicolumn{5}{c}{NuScenes}                                                   \\
\multicolumn{1}{l|}{}                                                            &                                                                                           & Car           & Ped           & Van           & Cyclist       & Mean          & Car           & Truck         & Trailer       & Bus           & Mean           \\
\multicolumn{1}{l|}{}                                                            &                                                                                           & 6424          & 6088          & 1248          & 308           & 14068         & 32302         & 8646          & 2297          & 2215          & 45460          \\ 
\hline\hline
\multirow{9}{*}{\rotatebox{90}{Success (\%) }}                                                                & SC3D~\cite{giancola2019leveraging}                                                                                     & 41.3          & 18.2          & 40.4          & 41.5          & 31.2          & 30.6          & 23.5          & 27.4          & 23.6          & 28.7           \\
                                                                                 & SC3D-RPN~\cite{zarzar2019efficient}                                                                                  & 36.3          & 17.9          & -             & 43.2          & -             & -             & -             & -             & -             & -              \\
                                                                                 & FSiamese~\cite{zou2020f}                                                                                  & 37.1          & 16.2          & -             & 47.0          & -             & -             & -             & -             & -             & -              \\
                                                                                 & 3DSiamRPN~\cite{fang20203d}                                                                                 & 58.2          & 35.2          & 45.6          & 36.1          & 46.6          & -             & -             & -             & -             & -              \\
                                                                                 & P2B~\cite{qi2020p2b}                                                                                       & 56.2          & 28.7          & 40.8          & 32.1          & 42.4          & 34.6          & 25.2          & 30.0          & 28.4          & 32.3           \\
                                                                                 & MLVSNet~\cite{wang2021mlvsnet}                                                                                   & 56.0          & 34.1          & 52.0          & 34.3          & 45.7          & -             & -             & -             & -             & -              \\
                                                                                 & PTT~\cite{shan2021ptt}                                                                                       & \textbf{67.8} & 44.9          & 43.6          & 37.2          & \textbf{55.1} & -             & -             & -             & -             & -              \\
                                                                                 & BAT~\cite{zheng2021box}                                                                                       & 60.5          & 42.1          & 52.4          & 33.7          & 51.2          & 36.8          & 28.6          & 31.8          & 30.2          & 34.7           \\
                                                                                 & DMT (Ours)                                                                              & 66.4          & \textbf{48.1} & \textbf{53.3} & \textbf{70.4} & \textbf{55.1} & \textbf{43.8} & \textbf{51.3} & \textbf{46.8} & \textbf{38.2} & \textbf{44.0}  \\ 
\hline\hline
\multirow{9}{*}{\rotatebox{90}{Precision (\%)}}                                                                & SC3D~\cite{giancola2019leveraging}                                                                                      & 57.9          & 37.8          & 47.0          & 70.4          & 48.5          & 35.9          & 24.8          & 24.8          & 21.8          & 32.5           \\
                                                                                 & SC3D-RPN~\cite{zarzar2019efficient}                                                                                  & 51.0          & 47.8          & -             & 81.2          & -             & -             & -             & -             & -             & -              \\
                                                                                 & FSiamese~\cite{zou2020f}                                                                                  & 50.6          & 32.2          & -             & 77.2          & -             & -             & -             & -             & -             & -              \\
                                                                                 & 3DSiamRPN~\cite{fang20203d}                                                                                 & 76.2          & 56.2          & 52.8          & 49.0          & 64.9          & -             & -             & -             & -             &                \\
                                                                                 & P2B~\cite{qi2020p2b}                                                                                       & 72.8          & 49.6          & 48.4          & 44.7          & 60.0          & 37.6          & 25.2          & 26.7          & 27.6          & 34.2           \\
                                                                                 & MLVSNet~\cite{wang2021mlvsnet}                                                                                   & 74.0          & 61.1          & 61.4          & 44.5          & 66.6          & -             & -             & -             & -             & -              \\
                                                                                 & PTT~\cite{shan2021ptt}                                                                                       & \textbf{81.8} & 72.0          & 52.5          & 47.3          & 74.2          & -             & -             & -             & -             & -              \\
                                                                                 & BAT~\cite{zheng2021box}                                                                                       & 77.7          & 70.1          & \textbf{67.0} & 45.4          & 72.8          & 39.5          & 28.4          & 30.5          & 29.5          & 36.4           \\
                                                                                 & DMT (Ours)                                                                               & 79.4          & \textbf{77.9} & 65.6          & \textbf{92.6} & \textbf{75.8} & \textbf{48.3} & \textbf{51.1} & \textbf{40.3} & \textbf{31.9} & \textbf{47.3}  \\
\hline
\end{tabular}}
\end{table*}

\begin{figure*}[t!]
\vspace{-0.3cm}
\centering
\includegraphics[width=0.9\linewidth]{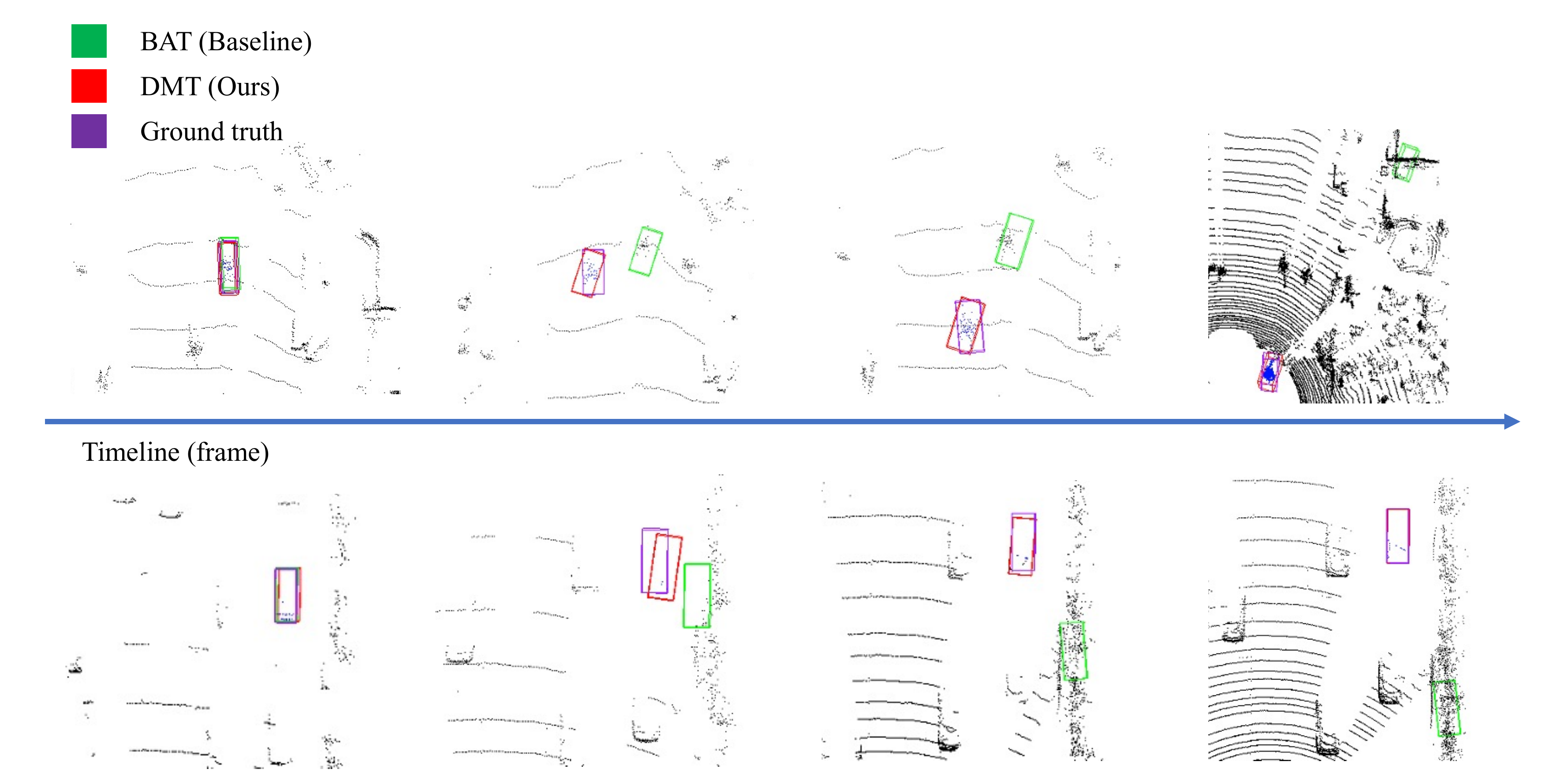}   
\caption{Visualizations of the example results of DMT compared with BAT.
The point clouds of tracked objects are shown in blue. (Top) shows the results for test instances from the Cyclist category. There are two cyclists nearby, and our DMT can maintain the correct track while BAT drifts to the wrong object. (BOTTOM) shows the results for test instances from the Car category. Although the point clouds are extremely sparse ($< 10$ points), our DMT still tracks the object.}
\label{fig:visualization} 
\end{figure*}

{{\bf Comparisons on KITTI}. Following \cite{qi2020p2b, zheng2021box}, we generate the search area centered on the previous result in the inference stage to meet the requirement of real scenarios.
The results in Table \ref{tab: experimental results} show that the proposed DMT outperforms other 3D trackers significantly. Specifically, 
when we mix all categories together to test the average performance following previous trackers, our average performance is 55.1, outperforming BAT by $\sim$4\% on Success, indicating the effectiveness of the  proposed DMT. When compared with PTT for the rigid object (e.g., Van) tracking, DMT has a significant advantage ($\sim$10\% ) over PTT in the less-frequent van category \jimmy{in terms of the success metric}. However, DMT does not achieve the highest performance in the more-frequent Car category. The transformer-based tracker PTT can learn better features of rigid objects since it has complex network architectures and more parameters but relies on more data to train the networks. 
Qualitative results are given in Section \ref{sec: Results visualization}. 

To demonstrate its generalizability for non-rigid object tracking, we compare it with other trackers on Pedestrian and Cyclist. For Pedestrian, we observe that DMT outperforms BAT and PTT by $\sim$8\% and $\sim$6\% on Precision respectively, indicating the effectiveness of our tracking pipeline. 
Amazingly, DMT outperforms BAT and PTT by a large margin for the cyclist category, achieving about $\sim$47\%/$\sim$45\% improvement for Precision. This phenomenon can be explained as follows: 1) The amount of training and testing samples is extremely small; 2) Our method DMT is less sensitive to interference with non-rigid objects in the search area; 3) DMT is simple yet effective, thus relying on less data to train better networks. The visualized results are shown in Fig. \ref{fig:visualization}. This also demonstrates that our method can achieve better performance, especially when having less data compared with BAT. }

{\bf Comparisons on NuScenes.} For the Car category, DMT achieves the best performance of 43.8/48.3 for Success/Precision, exceeding the performance of the current state-of-the-art method BAT \cite{zheng2021box} by $\sim$7\%/$\sim$9\%, respectively. Notably, for the Truck and Trailer categories, DMT achieves $\sim$23\% and $\sim$20\% improvements over BAT for Precision, which demonstrates that our motion-guided pipeline is more effective, especially on the more challenging dataset. Moreover, for the Bus category, which has the fewest training samples, our DMT still outperforms BAT by a large margin of $8\%$ in terms of the Success metric. Compared with the baseline method BAT, the performance of our DMT shows significant improvements ($\sim$10\% on average) in terms of all categories. Note that PTT/MLVSNet did not present results on NuScenes in their papers.

\subsection{Computational cost analysis}
\label{sec: time analysis}
In this section, we analyze the required computational resources of different 3D trackers in terms of the number of parameters, floating point operations (FLOPs), and running speed.
For a fair comparison, here we test our method on all KITTI-Car frames with a single NVIDIA RTX3090 GPU. As shown in Table \ref{tab: time results} and Fig.~\ref{fig:cover} (Left), our method uses less time per frame with fewer FLOPs compared with other trackers. Notably, despite the fact that our network includes an LSTM model, the number of parameters in our model are the same as P2B, while our model is significantly faster (57\% improvement in FPS) and simpler (36\% improvement in FLOPs) using the same RTX3090 GPU. In addition, the running speed of MLVSNet is close to ours. However, our DMT is lighter (i.e., with fewer model parameters)  and can achieve much better performance on the KITTI dataset (see Table \ref{tab: experimental results}), demonstrating that our method is simple yet effective.
\begin{table}[ht!]
\caption{Computational cost requirements of different 3D single object trackers on KITTI-Car. * indicates the FPS is taken from the corresponding paper.}
\centering
\resizebox{0.47\textwidth}{!}{
\begin{tabular}{cccccc} 
\toprule
Method       & \begin{tabular}[c]{@{}c@{}}Modality \end{tabular}  & \begin{tabular}[c]{@{}c@{}}Params \end{tabular} & \begin{tabular}[c]{@{}c@{}}FLOPs \end{tabular} & \begin{tabular}[c]{@{}c@{}}FPS \end{tabular}  &  \begin{tabular}[c]{@{}c@{}}Platform \end{tabular}                                                   \\ 
\hline
\hline
SC3D\cite{giancola2019leveraging}         & LiDAR     &   -     &  -     & 1.8* & 1080Ti                                                    \\
FSiamese\cite{zou2020f}     & LiDAR+RGB &   -     &    -   & 4.9* &1080Ti                                                       \\
3DSiamRPN\cite{fang20203d}    & LiDAR     &   -      &   -    & 20.8* & 1080Ti                                                    \\
P2B\cite{qi2020p2b}          & LiDAR     & \textbf{5.4M}       & 4.65G     & 45.5 & 3090                                                    \\
MLVSNet\cite{wang2021mlvsnet}          & LiDAR     & 7.6M       & -      & 70.0* & 1080Ti                                                     \\
PTT\cite{shan2021ptt}          & LiDAR     & -       & -      & 45  & 3090
                      \\
BAT\cite{zheng2021box}          & LiDAR     & 5.9M       & 3.05G      & 68.0 &  3090
                      \\
DMT (Ours)  & LiDAR     & \textbf{5.4M}       & \textbf{2.98G}      & \textbf{71.5} & 3090 \\
\bottomrule
\end{tabular}}
\label{tab: time results}
\end{table}

\subsection{Results visualization} \label{sec: Results visualization}
According to the different categories and difficulties of the targets, we select and visualize some advantageous cases of our DMT in Fig.~\ref{fig:visualization}. 
Four frames sorted by time from a full sequence are selected from the Cyclist and Car categories, respectively. For the cyclist target, the point clouds of the target and the tracked results are shown in the top of Fig.~\ref{fig:visualization}. In this example, BAT tracks the cyclist wrongly when there are two similar cyclists in the surrounding area. Our method can track the target accurately and tightly, indicating our method is more robust in complex scenarios.
Furthermore, we display the tracked results in the Car category, which is shown in the bottom of Fig.~\ref{fig:visualization}. Here, BAT fails in the extremely sparse scenes (fewer than 10 points), but our DMT works well, which  shows that our proposed method can indeed cope with point sparsity.

\section{Discussions} \label{section: Discussion}
In this section, we analyze the effectiveness of the important modules in our DMT, including both the motion prediction module (MPM) and the explicit voting module (EVM). We also discuss the choices of MPM, template generation strategies, sampling distances for training the EVM, the number of sampled training points, and the robustness to object motion patterns.

\subsection{Ablation studies of DMT components}\label{subsection: Ablation study}

\begin{table}[ht]
\centering
\caption{Ablation studies of motion prediction module (MPM) and explicit voting module (EVM) on KITTI-Car.}
\resizebox{0.47\textwidth}{!}{
\begin{tabular}{ccccc} 
\toprule
Method       & MPM & EVM & Success & Precision  \\ 
\hline\hline
BAT\cite{zheng2021box}          &          &          & 60.5    & 77.7       \\
DMT\_MP    & \checkmark         &          & -       & 37.0      \\
DMT\_EV    &          & \checkmark         & 54.0   & 64.1      \\
DMT & \checkmark         & \checkmark         & \textbf{66.4}    & \textbf{79.4}       \\
\bottomrule
\end{tabular}}
\label{tab:results}
\end{table}

We first conduct an ablation study on the necessity of the EVM and MPM.
All studies are conducted on KITTI-Car.
We remove the EVM and the MPM in our network one by one, which is denoted as DMT\_MP and DMT\_EV. Both variations have the same structure as DMT except for the removed module. The baseline model is BAT. 

The results are shown in Table~\ref{tab:results}.  We obtain four conclusions from these results. (1) The potential target center estimated by the \jimmy{MPM} is extremely inaccurate, only achieving 37\jimmy{\%} for Precision. Note that the \jimmy{MPM} in our network cannot regress the orientation of the target, and thus we cannot compute the Success value.
\xia{(2) The precision without the EVM is 37\jimmy{\%} (DMT\_MP), and with EVM is 79.4\jimmy{\%} (DMT), which demonstrates that EVM can estimate an effective target-specific point feature to further refine the prediction of the MPM. }
(3) Comparing DMT\_EV with BAT, the performance of DMT\_EV degrades about 6\% and 13\% in terms of Success and Precision, respectively. 
This is consistent with our expectation that we use a simpler explicit voting module, removing the complicated RPN module. 
(4) Our full pipeline achieves the best performance, which demonstrates the two modules are mutually beneficial and necessary. In addition, even if the \jimmy{MPM} provides inaccurate results, DMT achieves satisfactory performance due to the explicit voting module.

\subsection{The choice of motion prediction module}
\begin{figure}[t!]
\vspace{-0.3cm}
\centering
\includegraphics[width=1.0\linewidth]{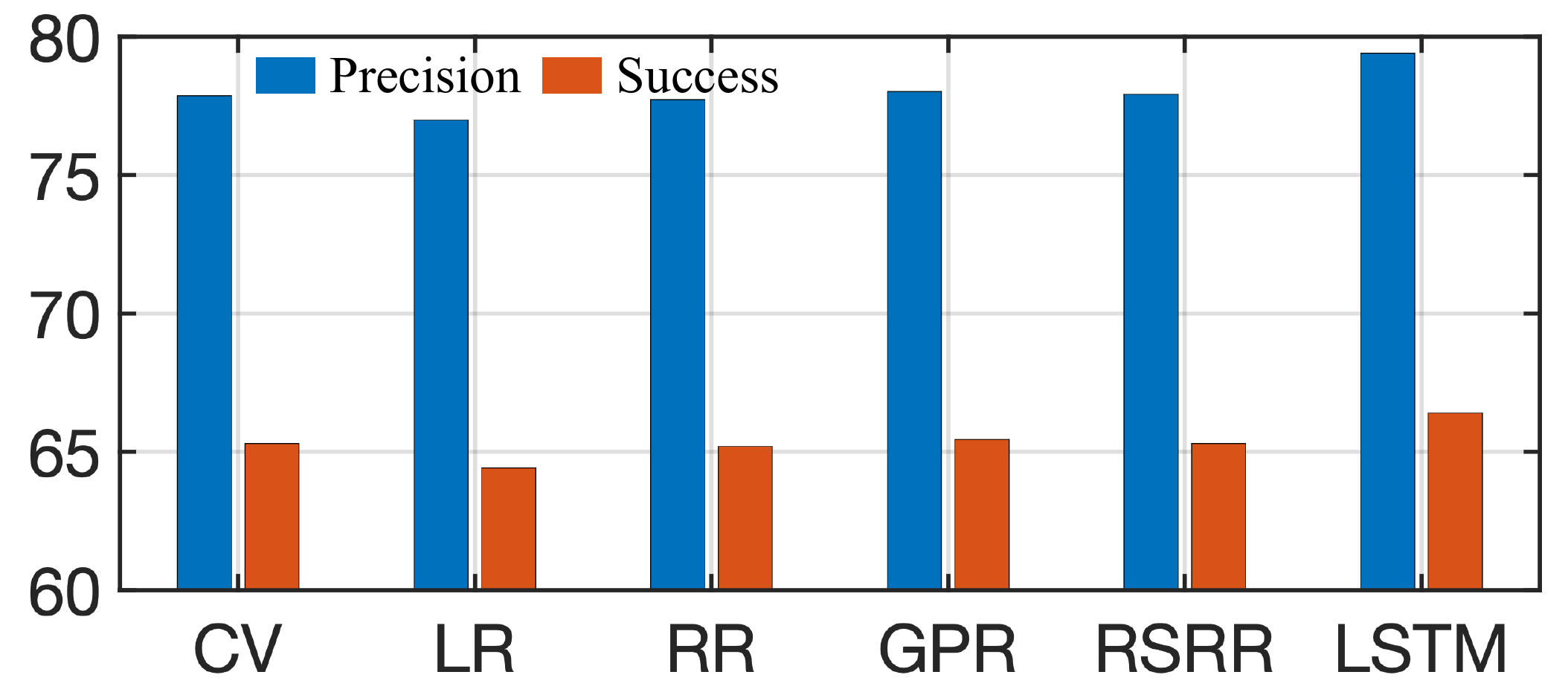}    
\vspace{-0.3cm}
\caption{Comparison of using various regression or prediction models as our motion prediction module on KITTI-Car.}
\label{fig:motion selection} 
\vspace{-0.3cm}
\end{figure}

In Fig.~\ref{fig:motion selection}, we compare various types of motion prediction models on \xia{KITTI-Car}. The compared models include constant velocity (CV), linear regression (LR), ridge regression (RR), Gaussian process regression (GPR), RANSAC with ridge regression, and LSTM models. The LR, RR, GPR, and RSRR models are trained in the same way as the LSTM model, i.e., using the same sampled tracklets from the training data in KITTI-Car for offline training. These models are then applied to motion prediction during online testing without further updating. 
In Fig.~\ref{fig:motion selection}, the differences between the various models are not significant, which implies that our DMT is not sensitive to the MPM selection. 
This is because our EVM is trained to predict GT bounding boxes from diverse sampled locations in the training stage, which makes it less sensitive to 
noisy predicted target center locations. The sequence-to-sequence prediction LSTM model achieves the best Precision (79.4\%) and Success (66.4\%) due to its better sequence modeling ability.

\subsection{Template generation strategy}


We next explore the performance of our DMT with four template generation strategies following \cite{zheng2021box}, including ``the first ground truth,'' ``the previous result,'' ``the first ground-truth and previous result,'' and ``all previous results.'' \revise{``The first ground truth'' generates a template using the target in the first frame (the ground truth). ``The previous result'' uses the result in the last frame predicted by the network, while ``all previous results'' concatenates the points in all previous results. To update the template efficiently, the default setting is ``the first ground truth and previous result'', which concatenates the target in the first frame with the prediction result in the last frame.}

Table \ref{tab: template results} shows the Success/Precision results with different settings for different trackers on KITTI-Car. \jimmy{Note that P2B, BAT, and DMT use the same PointNet++ backbone. The specific design in our DMT enables it to achieve better tracking performance than the other trackers under different template generation settings.} 
Specifically, DMT achieves the best performance when using the ``all previous'' strategy, outperforming BAT and P2B by large margins ($\sim$8\% and $\sim$12\%, respectively).
Another finding is that P2B, BAT and our DMT all report degraded results under the ``all previous'' setting since these trackers did not train the networks using all previous results for efficiency, while SC3D did. Despite this, the superior overall performance of DMT in Table \ref{tab: template results} suggests that DMT better utilizes motion cues from all previous predictions compared with BAT.

\begin{table}[ht!]
\vspace{-0.2cm}
\caption{Different strategies for template generation. 3D trackers are evaluated on KITTI-Car.}
\vspace{-0.5cm}
\centering
\resizebox{0.47\textwidth}{!}{
\begin{tabular}[t]{cccccc} 
\toprule
\multicolumn{1}{l}{}       & Method                                  & \begin{tabular}[c]{@{}c@{}}The First\\~GT\end{tabular} & \begin{tabular}[c]{@{}c@{}}Previous\\~Result\end{tabular} & \begin{tabular}[c]{@{}c@{}}First \& \\Previous\end{tabular} & \begin{tabular}[c]{@{}c@{}}All \\Previous~\end{tabular}  \\ 
\hline
\hline
\multirow{5}{*}{\rotatebox{90}{Success }}   & SC3D\cite{giancola2019leveraging}                                    & 31.6         & 25.7            & 34.9            & 41.3           \\
                           & P2B~\cite{qi2020p2b}                                     & 46.7         & 53.1            & 56.2            & 51.4           \\
                           & BAT~\cite{zheng2021box}                                     & 51.8         & 59.2            & 60.5            & 55.8           \\
                           & DMT (Ours) &\textbf{54.3}       & \textbf{63.8}           & \textbf{66.4}            & \textbf{63.5}        \\ 
\hline
\hline
\multirow{5}{*}{\rotatebox{90}{Precision}} & SC3D\cite{giancola2019leveraging}                                    & 44.4         & 35.1            & 49.8            & 57.9           \\
                           & P2B~\cite{qi2020p2b}                                     & 59.7         & 68.9            & 72.8            & 66.8           \\
                           & BAT~\cite{zheng2021box}                                     & 65.5         & 75.6            & 77.7            & 71.4           \\
                           & DMT (Ours)                      & \textbf{67.2}        & \textbf{76.7}           & \textbf{79.4}            & \textbf{75.9}          \\
\bottomrule
\end{tabular}}
\label{tab: template results}
\vspace{-0.4cm}
\end{table}

\subsection{Sampling distance for \jimmy{training EVM}}
\jimmy{In this section, we explore the network performance with different sampling distances (i.e., the distances between the sampled points and the ground-truth center) in the training of the EVM}.
As mentioned in Section \ref{sec: explicit voting module}, the distance should not be too large to maintain stable training. We conduct an ablation experiment on KITTI-Car, choosing the distance values from $0.65$ to $0.95$. As shown in Table \ref{tab: distance results}, the performance of DMT reaches its peak with a distance value of $0.75$. When the distance expands to $0.95$, the performance steadily degrades. This implies the distances between the sampled points and the ground-truth center are still a little large so some outliers are picked. On the other hand, the network performance drops when the distance is set to $0.65$. Thus, in this paper, we fix the values to $0.75$ for the best performance.

\begin{table}[ht]
\vspace{-0.2cm}
\centering
\caption{Sampling distance analysis for DMT. We evaluate DMT on KITTI-Car.}
\vspace{-0.2cm}
\resizebox{0.4\textwidth}{!}{
\label{tab: results}
\begin{tabular}{ccc} 
\toprule
Distance(m)~ & Success(\%) & Precision(\%)  \\ 
\hline\hline
0.65      & 64.0    & 77.0       \\
0.75      & \textbf{66.4}    & \textbf{79.4}       \\
0.85      & 63.0    & 77.5       \\
0.95      & 63.0    & 76.8       \\
\bottomrule
\end{tabular}}
\label{tab: distance results}
\vspace{-0.4cm}
\end{table}

\subsection{Number of sampled training points}
In the practical implementation, we sample various points around the ground-truth target center to mimic motion predictions during the online tracking process. In this section, we study how the number of sampled points affects the final tracking performance. Specifically, we vary the number of sampled points and report the corresponding performance on KITTI-Car in Table \ref{tab: number points}. We find that sampling dense points (i.e., 64) leads to better performance because dense sampling provides more comprehensive cases for training a more robust EVM. We also notice that the performance is not saturated, implying that better performance can be obtained by sampling a larger number of points.
\abc{However, in our current experiments we are limited by the GPU memory size.}

\begin{table}[ht]
\caption{Sampling point number analysis for DMT. We evaluate DMT on KITTI-Car.}
\vspace{-0.2cm}
\centering
\resizebox{0.4\textwidth}{!}{
\label{tab: results}
\begin{tabular}{ccc} 
\toprule
Number~ & Success(\%) & Precision(\%)  \\ 
\hline\hline
8      & 61.1    & 75.0       \\
16      & 62.2    & 75.7       \\
32      & 64.5    & 78.0       \\
64      & \textbf{66.4}    & \textbf{79.4}       \\
\bottomrule
\end{tabular}}
\label{tab: number points}
\vspace{-0.4cm}
\end{table}


\subsection{Robustness test for object motion patterns}

\begin{figure}[ht!]
\centering
\includegraphics[width=1.0\linewidth]{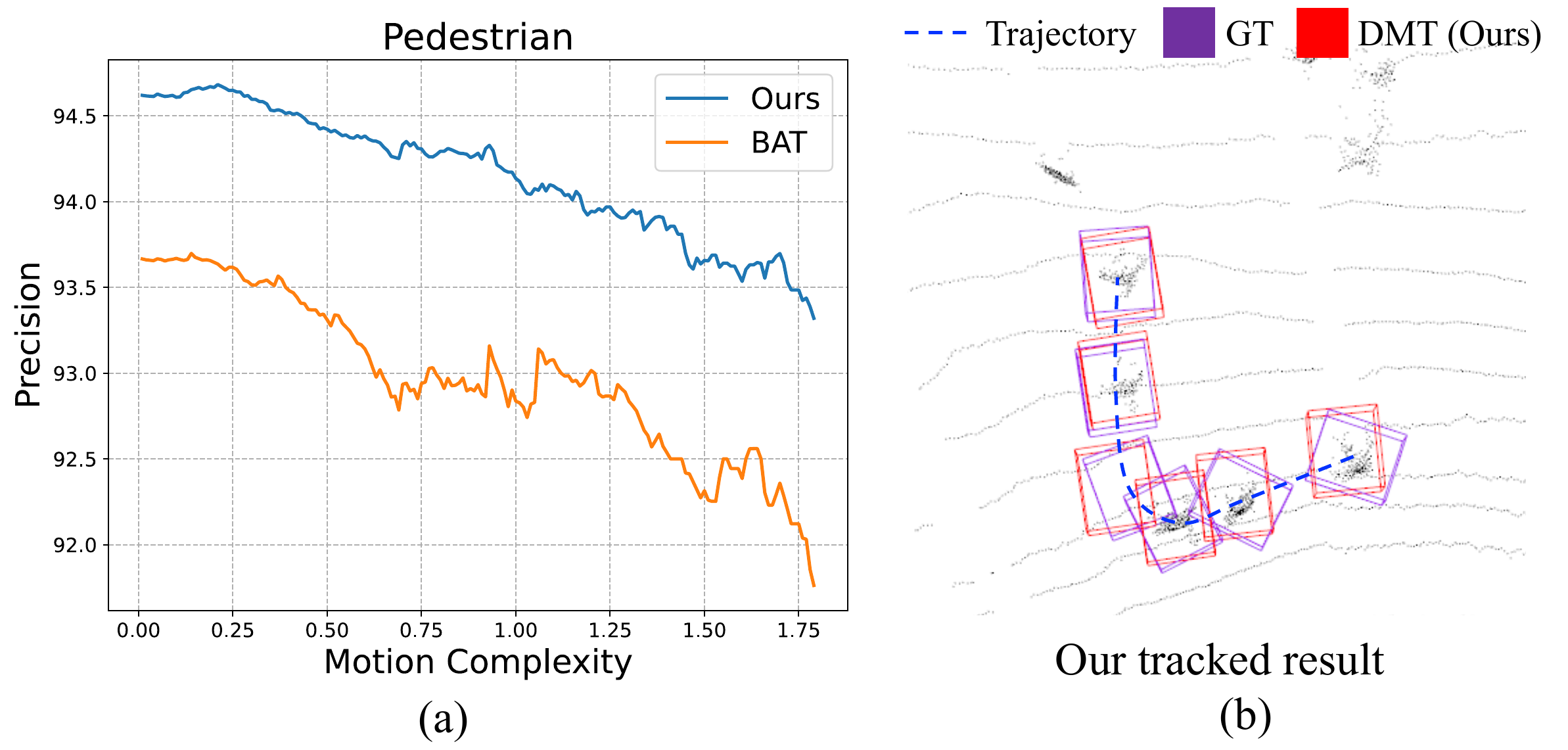}   
\caption{(a) Comparison of BAT and our DMT under various motion complexity on KITTI-Pedestrian. (b) Example results of DMT for complex motion patterns.}
\label{fig:visualization_pattern} 
\end{figure}

To better demonstrate the effectiveness of DMT on complex motion patterns, Fig.~\ref{fig:visualization_pattern}(a) shows the comparison of our DMT and BAT on tracklets with different motion complexities. Here, motion complexity is defined as the average error of a simple constant velocity model.
Our method still performs better than the RPN-based 3D tracker BAT when the motion complexity increases, which demonstrates the robustness of our method to complicated motion patterns. The reason is that we randomly sample diverse points when training the EVM, which makes our method more effectively handle various motion patterns. To further demonstrate the superiority clearly, we also visualize one tracklet of a pedestrian having a complex trajectory in Fig.~\ref{fig:visualization_pattern}(b). DMT can track the target accurately despite the complicated motion pattern.

\noindent

\section{Conclusion}
\label{sec: conclusion}
In this paper, we propose DMT, a novel lightweight and detector-free network for 3D single object tracking. We design a motion prediction module for predicting a potential target center, explicitly leveraging spatial-temporal correlations from previous frames to explore prior knowledge. In addition, we propose a simplified voting module to accurately regress the 3D box with the guidance of the potential target center.
Experiments show that our DMT method is lighter, faster, and simpler and improves the tracking performance over state-of-the-art methods significantly. According to discussions on experimental results, the explicit voting module based on a potential target center is an advantage of our method. We hope that our work will inspire more investigation into lightweight, detector-free 3D single-object trackers.

{\small
\bibliographystyle{ieee_fullname}
\bibliography{egbib}

\begin{thebibliography}{10}\itemsep=-1pt

\bibitem{achlioptas2018learning}
Panos Achlioptas, Olga Diamanti, Ioannis Mitliagkas, and Leonidas Guibas.
\newblock Learning representations and generative models for 3d point clouds.
\newblock In {\em International Conference on Machine Learning}, pages 40--49.
  PMLR, 2018.

\bibitem{asvadi20163d}
Alireza Asvadi, Pedro Girao, Paulo Peixoto, and Urbano Nunes.
\newblock 3d object tracking using rgb and lidar data.
\newblock In {\em IEEE International Conference on Intelligent Transportation
  Systems}, pages 1255--1260. IEEE, 2016.

\bibitem{bertinetto2016fully}
Luca Bertinetto, Jack Valmadre, Joao~F Henriques, Andrea Vedaldi, and Philip~HS
  Torr.
\newblock Fully-convolutional siamese networks for object tracking.
\newblock In {\em Proceedings of the European Conference on Computer Vision},
  pages 850--865. Springer, 2016.

\bibitem{SiamFC}
Luca Bertinetto, Jack Valmadre, Joao~F Henriques, Andrea Vedaldi, and Philip~HS
  Torr.
\newblock Fully-convolutional siamese networks for object tracking.
\newblock In {\em Proceedings of the European Conference on Computer Vision},
  pages 850--865. Springer, 2016.

\bibitem{bibi20163d}
Adel Bibi, Tianzhu Zhang, and Bernard Ghanem.
\newblock 3d part-based sparse tracker with automatic synchronization and
  registration.
\newblock In {\em Proceedings of the IEEE/CVF Conference on Computer Vision and
  Pattern Recognition}, pages 1439--1448, 2016.

\bibitem{caesar2020nuscenes}
Holger Caesar, Varun Bankiti, Alex~H Lang, Sourabh Vora, Venice~Erin Liong,
  Qiang Xu, Anush Krishnan, Yu Pan, Giancarlo Baldan, and Oscar Beijbom.
\newblock nuscenes: A multimodal dataset for autonomous driving.
\newblock In {\em Proceedings of the IEEE/CVF Conference on Computer Vision and
  Pattern Recognition}, pages 11621--11631, 2020.

\bibitem{comport2004robust}
Andrew~I Comport, {\'E}ric Marchand, and Fran{\c{c}}ois Chaumette.
\newblock Robust model-based tracking for robot vision.
\newblock In {\em IEEE/RSJ International Conference on Intelligent Robots and
  Systems}, volume~1, pages 692--697. IEEE, 2004.

\bibitem{cui2020point}
Yubo Cui, Zheng Fang, and Sifan Zhou.
\newblock Point siamese network for person tracking using 3d point clouds.
\newblock {\em Sensors}, 20(1):143, 2020.

\bibitem{fang20203d}
Zheng Fang, Sifan Zhou, Yubo Cui, and Sebastian Scherer.
\newblock 3d-siamrpn: An end-to-end learning method for real-time 3d single
  object tracking using raw point cloud.
\newblock {\em IEEE Sensors Journal}, 21(4):4995--5011, 2020.

\bibitem{geiger2012we}
Andreas Geiger, Philip Lenz, and Raquel Urtasun.
\newblock Are we ready for autonomous driving? the kitti vision benchmark
  suite.
\newblock In {\em Proceedings of the IEEE/CVF Conference on Computer Vision and
  Pattern Recognition}, pages 3354--3361. IEEE, 2012.

\bibitem{giancola2019leveraging}
Silvio Giancola, Jesus Zarzar, and Bernard Ghanem.
\newblock Leveraging shape completion for 3d siamese tracking.
\newblock In {\em Proceedings of the IEEE/CVF Conference on Computer Vision and
  Pattern Recognition}, pages 1359--1368, 2019.

\bibitem{dsiam}
Qing Guo, Wei Feng, Ce Zhou, Rui Huang, Liang Wan, and Song Wang.
\newblock Learning dynamic siamese network for visual object tracking.
\newblock In {\em Proceedings of the IEEE/CVF International Conference on
  Computer Vision}, pages 1763--1771, 2017.

\bibitem{resnet}
Kaiming He, Xiangyu Zhang, Shaoqing Ren, and Jian Sun.
\newblock Deep residual learning for image recognition.
\newblock In {\em Proceedings of the IEEE/CVF Conference on Computer Vision and
  Pattern Recognition}, pages 770--778, 2016.

\bibitem{hochreiter1997long}
Sepp Hochreiter and J{\"u}rgen Schmidhuber.
\newblock Long short-term memory.
\newblock {\em Neural Computation}, 9(8):1735--1780, 1997.

\bibitem{hu2019joint}
Hou-Ning Hu, Qi-Zhi Cai, Dequan Wang, Ji Lin, Min Sun, Philipp Krahenbuhl,
  Trevor Darrell, and Fisher Yu.
\newblock Joint monocular 3d vehicle detection and tracking.
\newblock In {\em Proceedings of the IEEE/CVF International Conference on
  Computer Vision}, pages 5390--5399, 2019.

\bibitem{jiayao2022real}
Shan Jiayao, Sifan Zhou, Yubo Cui, and Zheng Fang.
\newblock Real-time 3d single object tracking with transformer.
\newblock {\em IEEE Transactions on Multimedia}, 2022.

\bibitem{kart2018make}
Ugur Kart, Joni-Kristian Kamarainen, and Jiri Matas.
\newblock How to make an rgbd tracker?
\newblock In {\em Proceedings of the European Conference on Computer Vision
  (ECCV) Workshops}, pages 0--0, 2018.

\bibitem{kart2019object}
Ugur Kart, Alan Lukezic, Matej Kristan, Joni-Kristian Kamarainen, and Jiri
  Matas.
\newblock Object tracking by reconstruction with view-specific discriminative
  correlation filters.
\newblock In {\em Proceedings of the IEEE/CVF Conference on Computer Vision and
  Pattern Recognition}, pages 1339--1348, 2019.

\bibitem{kim2021discriminative}
Chanho Kim, Li Fuxin, Mazen Alotaibi, and James~M Rehg.
\newblock Discriminative appearance modeling with multi-track pooling for
  real-time multi-object tracking.
\newblock In {\em Proceedings of the IEEE/CVF Conference on Computer Vision and
  Pattern Recognition}, pages 9553--9562, 2021.

\bibitem{kiran2021deep}
B~Ravi Kiran, Ibrahim Sobh, Victor Talpaert, Patrick Mannion, Ahmad~A
  Al~Sallab, Senthil Yogamani, and Patrick P{\'e}rez.
\newblock Deep reinforcement learning for autonomous driving: A survey.
\newblock {\em IEEE Transactions on Intelligent Transportation Systems}, 2021.

\bibitem{SiamRPN_plus}
Bo Li, Wei Wu, Qiang Wang, Fangyi Zhang, Junliang Xing, and Junjie Yan.
\newblock Siamrpn++: Evolution of siamese visual tracking with very deep
  networks.
\newblock In {\em Proceedings of the IEEE/CVF Conference on Computer Vision and
  Pattern Recognition}, pages 4282--4291, 2019.

\bibitem{li2018high}
Bo Li, Junjie Yan, Wei Wu, Zheng Zhu, and Xiaolin Hu.
\newblock High performance visual tracking with siamese region proposal
  network.
\newblock In {\em Proceedings of the IEEE/CVF Conference on Computer Vision and
  Pattern Recognition}, pages 8971--8980, 2018.

\bibitem{siamrpn}
Bo Li, Junjie Yan, Wei Wu, Zheng Zhu, and Xiaolin Hu.
\newblock High performance visual tracking with siamese region proposal
  network.
\newblock In {\em Proceedings of the IEEE/CVF Conference on Computer Vision and
  Pattern Recognition}, pages 8971--8980, 2018.

\bibitem{liu2020object}
Yuan Liu, Ruoteng Li, Yu Cheng, Robby~T Tan, and Xiubao Sui.
\newblock Object tracking using spatio-temporal networks for future prediction
  location.
\newblock In {\em European Conference on Computer Vision}, pages 1--17.
  Springer, 2020.

\bibitem{luo2018fast}
Wenjie Luo, Bin Yang, and Raquel Urtasun.
\newblock Fast and furious: Real time end-to-end 3d detection, tracking and
  motion forecasting with a single convolutional net.
\newblock In {\em Proceedings of the IEEE/CVF Conference on Computer Vision and
  Pattern Recognition}, pages 3569--3577, 2018.

\bibitem{machida2012human}
Eiji Machida, Meifen Cao, Toshiyuki Murao, and Hiroshi Hashimoto.
\newblock Human motion tracking of mobile robot with kinect 3d sensor.
\newblock In {\em Proceedings of SICE Annual Conference}, pages 2207--2211.
  IEEE, 2012.

\bibitem{merrill2022symmetry}
Nathaniel Merrill, Yuliang Guo, Xingxing Zuo, Xinyu Huang, Stefan Leutenegger,
  Xi Peng, Liu Ren, and Guoquan Huang.
\newblock Symmetry and uncertainty-aware object slam for 6dof object pose
  estimation.
\newblock In {\em Proceedings of the IEEE/CVF Conference on Computer Vision and
  Pattern Recognition}, pages 14901--14910, 2022.

\bibitem{qi2019deep}
Charles~R Qi, Or Litany, Kaiming He, and Leonidas~J Guibas.
\newblock Deep hough voting for 3d object detection in point clouds.
\newblock In {\em Proceedings of the IEEE/CVF International Conference on
  Computer Vision}, pages 9277--9286, 2019.

\bibitem{qi2017pointnet++}
Charles~Ruizhongtai Qi, Li Yi, Hao Su, and Leonidas~J Guibas.
\newblock Pointnet++: Deep hierarchical feature learning on point sets in a
  metric space.
\newblock {\em Advances in Neural Information Processing Systems}, 30, 2017.

\bibitem{qi2020p2b}
Haozhe Qi, Chen Feng, Zhiguo Cao, Feng Zhao, and Yang Xiao.
\newblock P2b: Point-to-box network for 3d object tracking in point clouds.
\newblock In {\em Proceedings of the IEEE/CVF Conference on Computer Vision and
  Pattern Recognition}, pages 6329--6338, 2020.

\bibitem{ristic2003beyond}
Branko Ristic, Sanjeev Arulampalam, and Neil Gordon.
\newblock {\em Beyond the Kalman filter: Particle filters for tracking
  applications}.
\newblock Artech House, 2003.

\bibitem{shan2021ptt}
Jiayao Shan, Sifan Zhou, Zheng Fang, and Yubo Cui.
\newblock Ptt: Point-track-transformer module for 3d single object tracking in
  point clouds.
\newblock {\em arXiv preprint arXiv:2108.06455}, 2021.

\bibitem{stoiber2022srt3d}
Manuel Stoiber, Martin Pfanne, Klaus~H Strobl, Rudolph Triebel, and Alin
  Albu-Sch{\"a}ffer.
\newblock Srt3d: A sparse region-based 3d object tracking approach for the real
  world.
\newblock {\em International Journal of Computer Vision}, 130(4):1008--1030,
  2022.

\bibitem{stoiber2022iterative}
Manuel Stoiber, Martin Sundermeyer, and Rudolph Triebel.
\newblock Iterative corresponding geometry: Fusing region and depth for highly
  efficient 3d tracking of textureless objects.
\newblock In {\em Proceedings of the IEEE/CVF Conference on Computer Vision and
  Pattern Recognition}, pages 6855--6865, 2022.

\bibitem{sun2021you}
Jiaming Sun, Yiming Xie, Siyu Zhang, Linghao Chen, Guofeng Zhang, Hujun Bao,
  and Xiaowei Zhou.
\newblock You don't only look once: Constructing spatial-temporal memory for
  integrated 3d object detection and tracking.
\newblock In {\em Proceedings of the IEEE/CVF International Conference on
  Computer Vision}, pages 3185--3194, 2021.

\bibitem{wang2020motion}
Jianren Wang and Yihui He.
\newblock Motion prediction in visual object tracking.
\newblock In {\em IEEE/RSJ International Conference on Intelligent Robots and
  Systems}, pages 10374--10379. IEEE, 2020.

\bibitem{wang2021mlvsnet}
Zhoutao Wang, Qian Xie, Yu-Kun Lai, Jing Wu, Kun Long, and Jun Wang.
\newblock Mlvsnet: Multi-level voting siamese network for 3d visual tracking.
\newblock In {\em Proceedings of the IEEE/CVF International Conference on
  Computer Vision}, pages 3101--3110, 2021.

\bibitem{wu2013online}
Yi Wu, Jongwoo Lim, and Ming-Hsuan Yang.
\newblock Online object tracking: A benchmark.
\newblock In {\em Proceedings of the IEEE/CVF Conference on Computer Vision and
  Pattern Recognition}, pages 2411--2418, 2013.

\bibitem{xia2021asfm}
Yaqi Xia, Yan Xia, Wei Li, Rui Song, Kailang Cao, and Uwe Stilla.
\newblock Asfm-net: Asymmetrical siamese feature matching network for point
  completion.
\newblock In {\em Proceedings of the ACM International Conference on
  Multimedia}, pages 1938--1947, 2021.

\bibitem{xia2021soe}
Yan Xia, Yusheng Xu, Shuang Li, Rui Wang, Juan Du, Daniel Cremers, and Uwe
  Stilla.
\newblock Soe-net: A self-attention and orientation encoding network for point
  cloud based place recognition.
\newblock In {\em Proceedings of the IEEE/CVF Conference on Computer Vision and
  Pattern Recognition}, pages 11348--11357, 2021.

\bibitem{xia2021vpc}
Yan Xia, Yusheng Xu, Cheng Wang, and Uwe Stilla.
\newblock Vpc-net: Completion of 3d vehicles from mls point clouds.
\newblock {\em ISPRS Journal of Photogrammetry and Remote Sensing},
  174:166--181, 2021.

\bibitem{memtrack}
Tianyu Yang and Antoni~B Chan.
\newblock Learning dynamic memory networks for object tracking.
\newblock In {\em Proceedings of the European Conference on Computer Vision},
  pages 152--167, 2018.

\bibitem{zarzar2019efficient}
Jesus Zarzar, Silvio Giancola, and Bernard Ghanem.
\newblock Efficient bird eye view proposals for 3d siamese tracking.
\newblock {\em arXiv preprint arXiv:1903.10168}, 2019.

\bibitem{updatenet}
Lichao Zhang, Abel Gonzalez-Garcia, Joost van~de Weijer, Martin Danelljan, and
  Fahad~Shahbaz Khan.
\newblock Learning the model update for siamese trackers.
\newblock In {\em Proceedings of the IEEE/CVF International Conference on
  Computer Vision}, pages 4010--4019, 2019.

\bibitem{siamdw}
Zhipeng Zhang and Houwen Peng.
\newblock Deeper and wider siamese networks for real-time visual tracking.
\newblock In {\em Proceedings of the IEEE/CVF Conference on Computer Vision and
  Pattern Recognition}, pages 4591--4600, 2019.

\bibitem{zheng2021box}
Chaoda Zheng, Xu Yan, Jiantao Gao, Weibing Zhao, Wei Zhang, Zhen Li, and
  Shuguang Cui.
\newblock Box-aware feature enhancement for single object tracking on point
  clouds.
\newblock In {\em Proceedings of the IEEE/CVF International Conference on
  Computer Vision}, pages 13199--13208, 2021.

\bibitem{zheng2018robust}
Feng Zheng, Ling Shao, and Junwei Han.
\newblock Robust and long-term object tracking with an application to vehicles.
\newblock {\em IEEE Transactions on Intelligent Transportation Systems},
  19(10):3387--3399, 2018.

\bibitem{zhou2020tracking}
Xingyi Zhou, Vladlen Koltun, and Philipp Kr{\"a}henb{\"u}hl.
\newblock Tracking objects as points.
\newblock In {\em Proceedings of the European Conference on Computer Vision},
  pages 474--490. Springer, 2020.

\bibitem{zou2020f}
Hao Zou, Jinhao Cui, Xin Kong, Chujuan Zhang, Yong Liu, Feng Wen, and Wanlong
  Li.
\newblock F-siamese tracker: A frustum-based double siamese network for 3d
  single object tracking.
\newblock In {\em IEEE/RSJ International Conference on Intelligent Robots and
  Systems}, pages 8133--8139. IEEE, 2020.

\end{thebibliography}
}
\end{document}